\documentclass[journal]{IEEEtran}
\usepackage{graphicx,float,stfloats,amsthm,amssymb,multirow,pifont,amsmath,scalerel,makecell}
\usepackage{bm,hyperref,xcolor,enumitem,booktabs,arydshln,footnote,url}
\usepackage[caption=false]{subfig}
\usepackage{color}
\usepackage{mathrsfs}
\usepackage{algorithm,algorithmic}
\usepackage{threeparttable}

\usepackage[subfigure,titles]{tocloft}
\usepackage[comma, numbers]{natbib}
\hypersetup{hypertex=true,
	colorlinks=true,
	linkcolor=blue,
	anchorcolor=blue,
	citecolor=blue}

\begin{document}
\title{Decomposition-based Unsupervised Domain Adaptation for Remote Sensing Image Semantic Segmentation}
\author{Xianping~Ma, Xiaokang~Zhang,~\IEEEmembership{Senior Member,~IEEE}, Xingchen~Ding, Man-On~Pun,~\IEEEmembership{Senior Member,~IEEE}, and~Siwei~Ma,~\IEEEmembership{Fellow,~IEEE,}
\thanks{This work was supported in part by the Guangdong Provincial Key Laboratory of Future Networks of Intelligence under Grant 2022B1212010001, and in part by the National Natural Science Foundation of China under Grant 42371374 and Grant 41801323. \textit{(Corresponding authors: Man-On Pun and Xiaokang~Zhang)}}
\thanks{Xianping~Ma and Man-On Pun are with the School of Science and Engineering, The Chinese University of Hong Kong, Shenzhen, Shenzhen 518172, China (e-mail:xianpingma@link.cuhk.edu.cn; SimonPun@cuhk.edu.cn).}
\thanks{Xiaokang~Zhang is with the School of Information Science and Engineering, Wuhan University of Science and Technology, Wuhan 430081, China (e-mail: natezhangxk@gmail.com).}
\thanks{Xingchen~Ding is with the School of Cyber Science and Engineering, Wuhan University, Wuhan 430072, China (e-mail: xingchen.ding@student.uts.edu.au).}
\thanks{Siwei~Ma is with the School of Electronic Engineering and Computer Science, Peking University, Beijing 100871, China (e-mail: swma@pku.edu.cn).}}
\maketitle

\begin{abstract}
Unsupervised domain adaptation (UDA) techniques are vital for semantic segmentation in geosciences, effectively utilizing remote sensing imagery across diverse domains.  However, most existing UDA methods, which focus on domain alignment at the high-level feature space, struggle to simultaneously retain local spatial details and global contextual semantics. To overcome these challenges, a novel decomposition scheme is proposed to guide domain-invariant representation learning. Specifically, multiscale high/low-frequency decomposition (HLFD) modules are proposed to decompose feature maps into high- and low-frequency components across different subspaces. This decomposition is integrated into a fully global-local generative adversarial network (GLGAN) that incorporates global-local transformer blocks (GLTBs) to enhance the alignment of decomposed features. By integrating the HLFD scheme and the GLGAN, a novel decomposition-based UDA framework called De-GLGAN is developed to improve the cross-domain transferability and generalization capability of semantic segmentation models. Extensive experiments on two UDA benchmarks, namely ISPRS Potsdam and Vaihingen, and LoveDA Rural and Urban, demonstrate the effectiveness and superiority of the proposed approach over existing state-of-the-art UDA methods. The source code for this work is accessible at \href{https://github.com/sstary/SSRS}{https://github.com/sstary/SSRS}.
\end{abstract}


\IEEEpeerreviewmaketitle

\section{Introduction}\label{sec:int}
Driven by the rapid development of Earth Observation (EO) technology, semantic segmentation has become one of the most important tasks in practical remote sensing applications, such as land cover monitoring, land planning, and natural disaster assessment \cite{hong2021more, 9756442}. In the era of deep learning, semantic segmentation models are normally trained by exploiting labeled datasets in a fully supervised manner \citep{shan2022class, 10538308, ma2024sam, ma2024rs3mamba}. However, their impressive performance is limited by the availability of high-quality annotated samples generated by the costly and laborious labeling process \citep{li2021learning, yan2023domain}. Furthermore, these supervised methods show poor generalization when the trained models are applied to data in other domains or unseen scenarios due to discrepancies between heterogeneous domains \citep{zhang2023cross}.  To circumvent the domain shift, the {\em Unsupervised} Domain Adaption (UDA) approach has been investigated, which aims to transfer knowledge from the source domain and achieve the desired semantic segmentation performance in the target domain \citep{peng2022domain,tsai2018learning, vu2019advent}.

Generally speaking, the core idea of UDA is to learn domain-invariant features through domain alignment based on optimization principles such as discrepancy metrics \citep{liu2020multikernel} and adversarial learning \citep{peng2021full, huang2022two, hong2023cross, huang2023cross}. In particular, the adversarial-based methods were developed upon the generative adversarial network (GAN) that alternately trains a generator and a discriminator for domain alignment. In semantic segmentation tasks, adversarial-based methods attempt to align two domains at image-level \citep{tasar2020colormapgan, li2022stepwise}, feature-level \citep{zhang2020domain, mbatagan} or output-level \citep{ni2023category, chen2023memory, huang2024adversarial, ning2024domain}. Since global context modeling is important for remote sensing vision tasks, transformer \citep{vaswani2017attention}, with its superiority by its ability to model long-range contextual information, has facilitated many GAN-based UDA methods by enhancing feature alignment \citep{li2022unsupervised, zhang2022unsupervised}.

\begin{figure}[h]
\centering
{\includegraphics[width=1\linewidth]{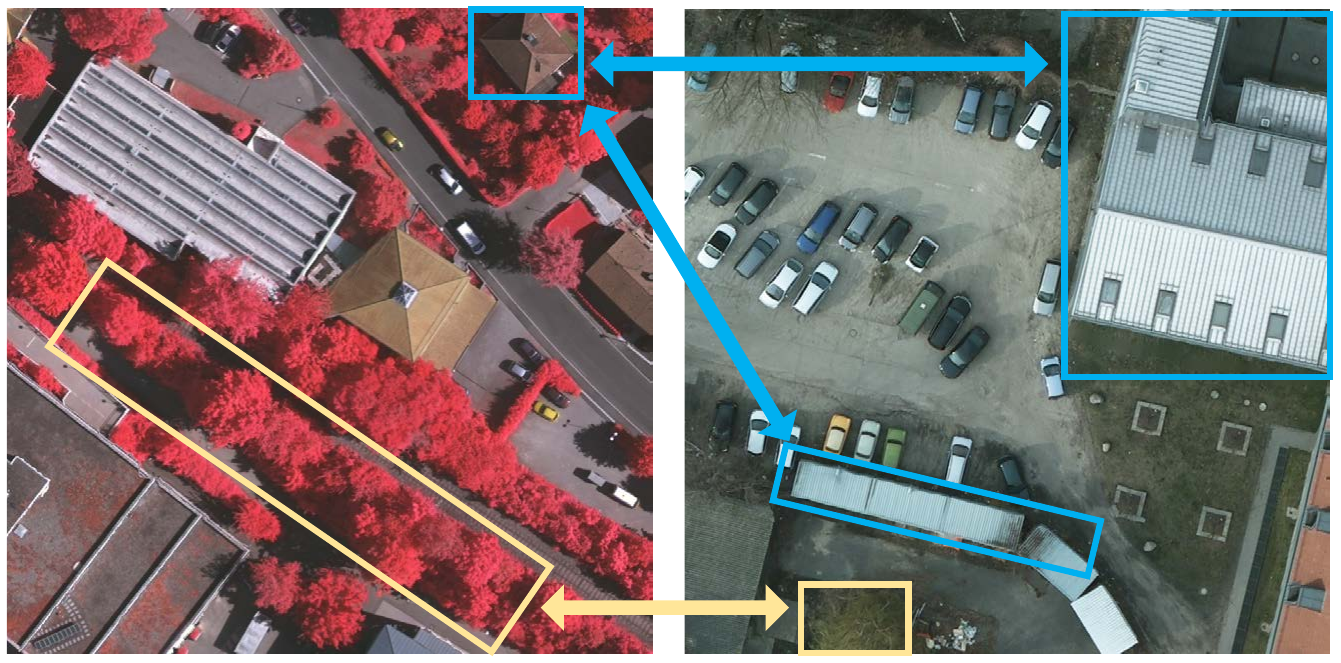}}
\caption{The main challenges of UDA semantic segmentation in remote sensing images. Larger-scale variations of trees and buildings are observed within and across different domains. For example, the building highlighted in the left figure is much smaller than those in the right figure, while the trees highlighted in the left figure are connected and much larger than that on the right. Additionally, the boundaries of ground objects on both domains often exhibit varying styles and complexities.}
\label{fig0}
\end{figure}

Notably, UDA on semantic segmentation of remote sensing images presents unique challenges. The ground objects and their spatial relationships are complex in fine-resolution remote sensing images \citep{sui2024diffusion}. Further, it becomes more evident when ground objects are collected from different locations. Larger-scale variations and more complicated boundaries can be observed as shown in Fig.~\ref{fig0}. Nevertheless, it has been also observed that cross-domain remote sensing images share certain similarities in both local and global contexts. Therefore, we argue that both {\em detailed local features} and {\em abstract global contexts} are crucial for domain alignment in semantic segmentation. The former includes the fine structure (e.g., corners and edges) \citep{liu2018semantic}, small-scale spatial variations and patterns, and precise layouts of parts of objects or complete small objects. In contrast, the latter encompasses types of landscapes, significant geographical features (e.g., mountains and rivers), and the spatial relationships between ground objects. However, existing UDA methods generally capture global contexts by aligning high-level features \cite{zhang2020domain, mbatagan, song2019domain} or learn local contexts by designing loss functions \citep{chen2022unsupervised, ma2023domain, gao2023prototype}. The latter exploits local constraints as an auxiliary way to learn global domain-invariant representations. Therefore, they struggle to simultaneously preserve and align local spatial details and global contextual semantics. From the perspective of model structure, existing methods primarily focus on the decoder and discriminator of the GAN framework, neglecting the encoder stage, which contains richer multiscale global and local domain-invariant semantic information. This oversight leads to a gap in knowledge transfer across domains.

To address these problems, we propose a decomposition-driven UDA method based on a global-local GAN model, namely De-GLGAN, considering alignment in low-frequency global representations and high-frequency local information. Firstly, we propose a high/low-frequency {\em decomposition} (HLFD) scheme that decomposes the feature map into high-frequency and low-frequency components through a CNN-based extractor and a window-based attention module, respectively. By aligning high- and low-frequency information simultaneously, the generator can learn more transferable features from the source domain. In particular, the HLFD scheme with a frequency alignment loss can be easily embedded into other UDA approaches to facilitate the learning of domain-invariant representations. To further exploit the decomposed information in an adversarial learning framework, a novel GAN model named GLGAN is proposed by exploiting a decoder and a global-local discriminator (GLDis), both of which are derived from the global-local transformer block (GLTB) \citep{wang2022unetformer}. With the exploration of global contexts and local spatial information, the former facilitates the generation of accurate segmentation maps, whereas the latter efficiently distinguishes the origins of segmentation maps. The contributions of this work are summarized in the following:
\begin{itemize}[leftmargin=*]
\item A novel decomposition-based adaptation scheme is proposed to learn high- and low-frequency domain-invariant features at multiple scales using a frequency alignment loss for model optimization. To the best of our knowledge, this is the first work that introduces the idea of frequency decomposition into UDA for remote sensing image semantic segmentation.

\item We implement the decomposition-based adaptation scheme into a fully global-local adversarial learning-based UDA framework. It facilitates domain alignment by capturing cross-domain dependency relationships at different levels while exploiting global-local context modeling between two domains.

\item Extensive experiments on two UDA benchmarks, i.e., ISPRS Potsdam and Vaihingen, and LoveDA Rural and Urban confirm that the proposed De-GLGAN is more effective in overcoming the domain shift problem in cross-domain semantic segmentation than existing UDA methods. 
\end{itemize}

The remainder of this paper is organized as follows. We first review the related works in Sec.~\ref{sec:rel} while Sec.~\ref{sec:pro} introduces the proposed HLFD and GLGAN in detail. After that, Sec.~\ref{sec:experiment} elaborates on the experiment setup, experimental results and related analysis. Finally, the conclusion is given in Sec.~\ref{sec:con}.

\section{Related Work}\label{sec:rel}
\subsection{Cross-Domain Semantic Segmentation in Remote Sensing} 
Leveraging encoder-decoder-based networks, numerous methods of semantic segmentation have emerged in the field of remote sensing \citep{kotaridis2021remote}. Most existing methods are developed based on CNNs with residual learning and attention mechanisms to boost the representation learning \citep{diakogiannis2020resunet}. To address the limitations of CNNs in modeling global representations, many transformer-based methods \citep{ma2022crossmodal, ma2024multilevel} have been integrated under the encoder-decoder framework, leveraging the global context modeling capability of the self-attention mechanism \citep{vaswani2017attention, liu2021swin}. Recently developed remote sensing foundation models attempt to achieve domain-generalized performance by leveraging large-scale datasets and vision models \cite{9844015,10490262,wang2022advancing}. For downstream tasks on specific remote sensing datasets, transfer learning or fine-tuning is required.

In contrast to the straightforward semantic segmentation task, cross-domain semantic segmentation focuses on transferring knowledge from the source domain to the target domain. According to the transfer strategy, existing cross-domain semantic segmentation models can be divided into two main categories, namely the self-training approach \citep{yan2019triplet, chen2023exchange} and the adversarial training approach \citep{peng2021full, mbatagan}. The former generates pseudo-labels as the supervision information of the target domain using a model trained on the labeled source domain \citep{cai2022iterdanet}. However, pseudo-labeling inevitably introduces noise into model training, typically requiring a pre-defined threshold to filter out low-confidence pseudo-labels. Alternatively, the adversarial training approach focuses on aligning distributions of two domains at the image feature or output levels. The key to this method lies in the stability of the training process. Recently, there has been an increase in methods that integrate self-training techniques with adversarial learning to address UDA challenges, using a multi-stage training strategy \citep{yan2019triplet, zhang2021stagewise, liang2023unsupervised}. This work aims to develop a versatile and efficient framework based on adversarial training. Compared to these methods, this work develops an effective one-stage UDA framework with promising performance based on adversarial training.

\subsection{GAN-based UDA in Remote Sensing}
GAN-based UDA algorithms have been proven effective in cross-domain semantic segmentation for remote sensing imagery. \citep{benjdira2019unsupervised} proposed the first GAN-based model designed for cross-domain semantic segmentation from aerial imagery.  On this basis, TriADA \citep{yan2019triplet} introduced a triplet branch to improve the CNN-based feature extraction by simultaneously aligning features from the source and target domains. Recently, a self-training approach that employs high-confidence pseudo-labels was applied to incorporate category information \citep{ liang2023unsupervised, ma2023domain, shen2024optimal}. Furthermore, SADA \citep{song2019domain} and CS-DDA \citep{zhang2020domain} developed subspace alignment methods to constitute a shared subspace with high-level features. To improve the high-level feature extraction, attention mechanisms were introduced into GAN-based UDA. More specifically, CAGAN \citep{liu2022unsupervised} developed a covariance-based channel attention module to compute weighted feature maps, whereas CCAGAN \citep{chen2022unsupervised} used a category-certainty attention module to align the category-level features. MASNet \citep{zhu2023unsupervised} stored domain-invariant prototypical representations by employing a feature memory module while MBATA-GAN \citep{mbatagan} aligned high-level features by constructing a cross-attention-based transformer. Compared with the methods above, this work attempts to fully explore global-local context modeling in the generator and discriminator, as shown in Table~\ref{tab:gl_class}. Therefore, it can capture cross-domain global semantics and local detailed contexts while boosting the learning of domain-invariant features.

\begin{table}[h]
	\begin{center}
		\caption{GAN-based Model classification according to the utilization strategies of global and local information.}
		\label{tab:gl_class}
		\begin{tabular}{m{1.8cm}<{\centering}|m{1.6cm}<{\centering}|m{4.3cm}<{\centering}} 
			\hline
			\textbf{Generator} & \textbf{Discriminator} & \textbf{Method} \\\hline
			Local & Local &  HighDAN \citep{hong2023cross}, MBATA-GAN \citep{mbatagan}, CCAGAN \citep{chen2022unsupervised}, TriADA \citep{yan2019triplet}, GANAI \citep{benjdira2019unsupervised}, CAGAN \citep{liu2022unsupervised}, MASNet \citep{zhu2023unsupervised}\\\hline
			Global & Local & ALE-UDA \citep{zhang2022unsupervised}, MuGCDA \citep{xi2023multilevel}\\\hline
			Global & Global & TransGAN \citep{jiang2021transgan}, Swin-GAN \citep{wang2022swin}, HyperViTGAN \citep{he2022hypervitgan}, SWCGAN \citep{tu2022swcgan} \\\hline
			Global-Local & Global-Local & GLGAN (Ours) \\\hline
		\end{tabular}
	\end{center}
\end{table}

\subsection{Decomposition-based UDA in Remote Sensing}
The core idea of decomposition-based UDA is to project features into different subspaces before performing domain alignment in the corresponding subspaces. For instance, ToMF-B \citep{zhao2022unsupervised} decomposed the cross-domain features into task-related features and task-irrelevant features by analyzing the gradients of the predicted score corresponding to the labels. Furthermore, ST-DASegNet \citep{zhao2024self} introduced a domain disentangled module to extract cross-domain universal features and improve single-domain distinct features in a self-training framework whereas DSSFNet \citep{chen2024unsupervised} decoupled building features to learn domain-invariant semantic representations and domain-specific style information for building extraction. However, these techniques mainly concentrate on aligning high-level abstract features, overlooking the relationship between high/low-frequency details and the characteristics of ground objects. Therefore, it is crucial to address this shortfall by delving into frequency decomposition technology.

\section{Proposed Method}\label{sec:pro}
{In this section, we will first present the principle of frequency decomposition and introduce the HLFD module inspired by GLTB. After that, we propose the overall structure of De-GLGAN before elaborating on its generator and discriminator (GLDis). Finally, the loss functions used in training will be presented.}
\subsection{Preliminary: Frequency Decomposition}\label{sec:preli}
Before introducing our proposed UDA framework, we first revisit the basics of the multi-head self-attention (MHSA) design and the convolutional operation (Conv) from a frequency perspective. We employ the discrete Fourier transform (DFT) to analyze the feature maps regarding specific frequency components generated by MHSA and Conv, respectively. Let $\bm{X}\in \mathbb{R}^{H\times W}$ and $\bm{Y}\in \mathbb{R}^{C}$ represent an image in the spatial domain and its corresponding label, where $C$ denotes the number of classes. We transform $\bm{X}$ into the frequency spectrum using the DFT $\mathcal{F}$: $\mathbb{R}^{H\times W} \rightarrow \mathbb{C}^{H\times W}$ and revert the signals of the image from frequency back to the spatial domain using the inverse DFT $\mathcal{F}^{-1}$: $\mathbb{C}^{H\times W} \rightarrow \mathbb{R}^{H\times W}$. For a mask $m\in \{0,1\}^{H\times W}$, the low-pass filtering $\mathcal{M}^{S}_{\mathrm{low}}$ of size $S$ can be formally defined as \citep{bai2022improving}:
\begin{equation}
\mathcal{M}^{S}_{\mathrm{low}}=\mathcal{F}^{-1}\left(\bm{m}\odot\mathcal{F}(\bm{x})\right),
\end{equation}
where
\begin{equation}
m_{i,j}=
\begin{cases}
  1,~\mathrm{if~min}(|i-\frac{H}{2}|,|j-\frac{W}{2}|)\leq\frac{S}{2}\\
	0,~\mathrm{otherwise}
\end{cases},\label{eq:lf}
\end{equation}
where $\odot$ stands for the Hadamard product, and $m_{i,j}$ denotes the value of $m$ at position $(i,j)$.

Similarly, the high-pass filtering $\mathcal{M}^{S}_{\mathrm{high}}$ is defined as
\begin{equation}
\mathcal{M}^{S}_{\mathrm{high}}=\mathcal{F}^{-1}\left(\bm{m}\odot\mathcal{F}(\bm{x})\right), 
\end{equation}
where
\begin{equation}
m_{i,j}=
\begin{cases}
  0,~\mathrm{if~min}(|i-\frac{H}{2}|,|j-\frac{W}{2}|)\leq\frac{\mathrm{min}(H,W)-S}{2}\\
	1,~\mathrm{otherwise}
\end{cases}.\label{eq:hf}
\end{equation}
 According to the definitions in Eq.~\eqref{eq:lf} and Eq.~\eqref{eq:hf}, the low-frequency components are shifted to the center of the frequency spectrum while the high-frequency components are shifted away from the center of the frequency spectrum.

\begin{figure*}[t]
\centering
{\includegraphics[width=1.0\linewidth]{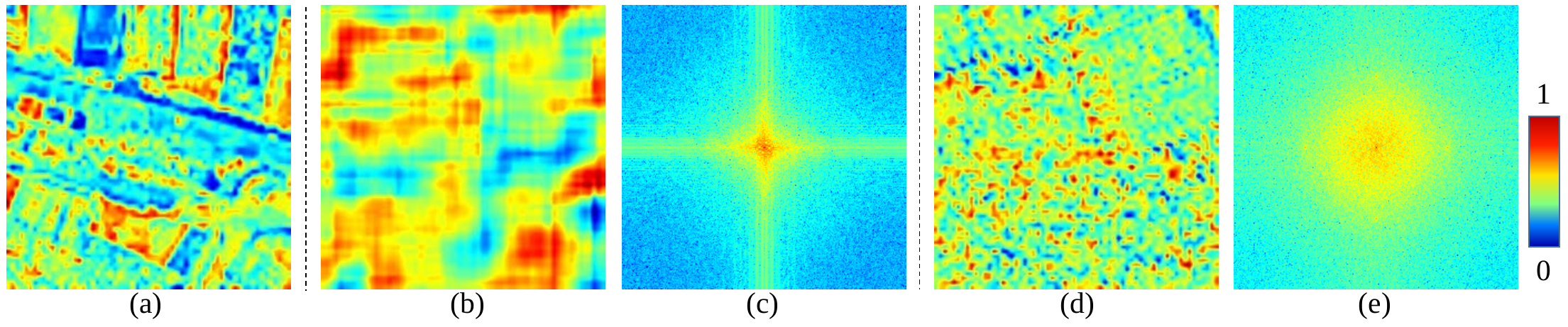}}
\caption{The basic ideas of our decomposition strategy. (a) Feature map extracted by the encoder, (b) feature map extracted further by MHSA and (c) its corresponding frequency spectrum; (d) feature map extracted further by Conv and (e) its corresponding frequency spectrum. The centroid represents low-frequency information, while the distance from it represents high-frequency information. More high-value points represent more information for the corresponding frequency.}
\label{fig1}
\end{figure*}

\begin{figure*}[h]
\centering
{\includegraphics[width=0.95\linewidth]{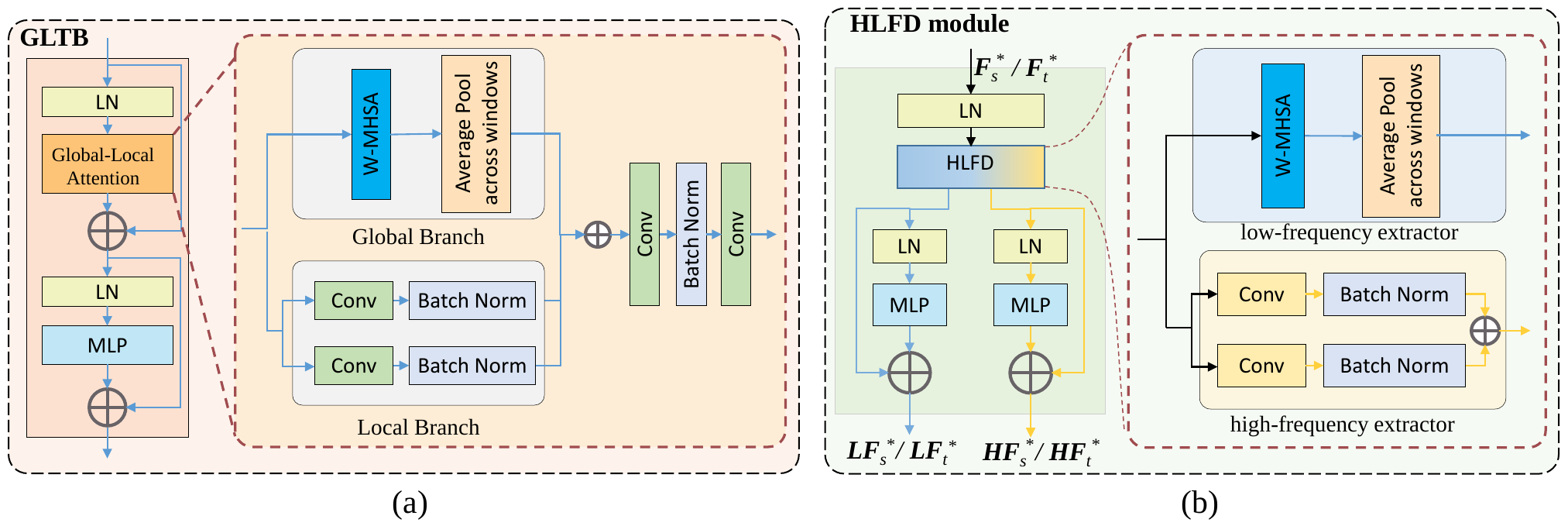}}
\caption{(a) GLTB propsoed in UNetformer \citep{wang2022unetformer} and (b) the proposed HLFD module. They have similar structures, both based on MHSA and Conv. However, the module structure and optimization goal of the network determine their different functions: one focuses on global-local contextual information extraction, while the other handles cross-domain feature decomposition.}
\label{fig2_1}
\end{figure*}

Fig.~\ref{fig1} illustrates the contrasting behaviors that MHSA and Conv demonstrate.  As shown in Fig.~\ref{fig1}(b) and (d), MHSA aggregates feature maps, whereas Conv disperses them. Furthermore, as depicted in Fig.~\ref{fig1}(c) and (e), Fourier analysis of feature maps reveals that MHSA focuses on low-frequency components, while Conv amplifies high-frequency components. In other words, MHSAs function as low-pass filters, while Conv serves as high-pass filters. Consequently, MHSA and Conv complement each other and can be jointly employed for frequency decomposition. More details can be found in \citep{wang2020high}.

Based on the discussions above, we propose to utilize the existing GLTB for the frequency decomposition. In particular, the global-local attention in GLTB consists of two parallel branches, namely the global and local branches, as presented in Fig.~\ref{fig2_1}(a). The former captures the global context between windows by exploiting the window-based multi-head self-attention (W-MHSA) \citep{liu2021swin} while maintaining the spatial consistency of ground objects through Average Pooling across windows, while the latter has a structure of two parallel convolutional layers with kernel sizes of $3$ and $1$, respectively. Each convolutional layer is followed by a batch normalization operation. The global-local attention leverages its two branches to investigate and incorporate global-local contextual information. Hereby, the outputs from the global and local branches are fused together before the resulting global-local context is further characterized by two convolution layers and a batch normalization operation.

It is observed that GLTB satisfies the requirements for frequency decomposition that needs a branch based on MHSA and a branch based on Conv. Therefore, we propose the HLFD module as depicted in Fig.~\ref{fig2_1}(b) by exploiting the GLTB structure to decompose multiscale features. However, in sharp contrast to GLTB, which is utilized for global-local information aggregation, the HLFD module utilizes two parallel branches for the decomposition. More specifically, a low-frequency extractor is designed to capture the low-frequency components by utilizing the W-MHSA while a high-frequency extractor is employed to extract high-frequency components using Conv operation. The output low- and high-frequency components are then processed by a LayerNorm (LN) layer and a multilayer perceptron (MLP), respectively. Finally, a novel model called De-GLGAN is established by capitalizing on the GLTB and HLFD modules.

\begin{figure*}[t]
\centering
{\includegraphics[width=0.9\linewidth]{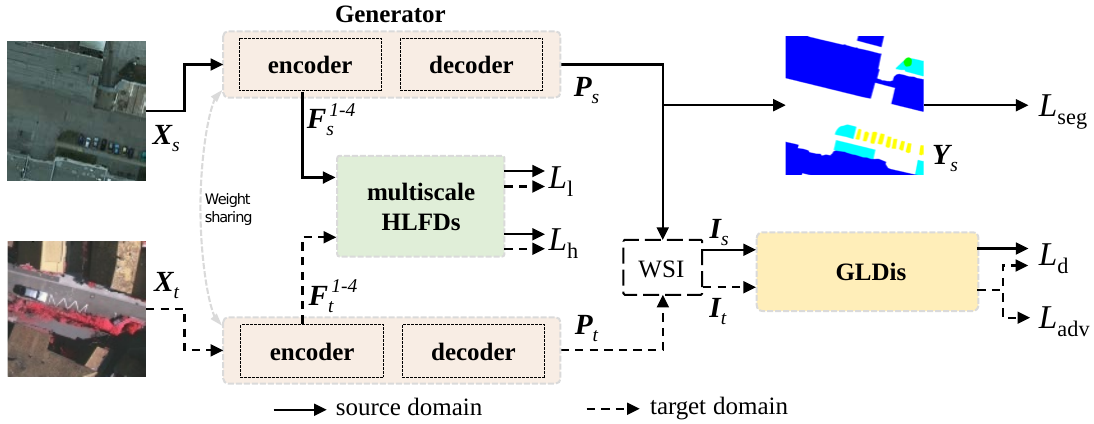}}
\caption{The overview of the proposed De-GLGAN, which is comprised of a Generator, a GLDis and the multiscale HLFDs. The Generator adopts the classic encoder-decoder structure, which can extract image features and predict pixel-wise category labels. The multiscale HLFDs are proposed to align cross-domain representations by decomposing multiscale features generated by the encoder. The GLDis further learns domain-invariant representations by adversarial learning strategy.}
\label{fig2}
\end{figure*}

\subsection{Pipeline}\label{overview}
Typical UDA tasks involve a labeled source dataset $D_{s}=\left\{(\bm{X}_{s}, \bm{Y}_{s})\right\}^{n_{s}}$ and an unlabeled target dataset $D_{t}=\left\{\bm{X}_{t}\right\}^{n_{t}}$ where $\bm{X}_{s}$ represents an image in the source domain with its corresponding label $\bm{Y}_{s}$ whereas $\bm{X}_t$ denotes the unlabeled image in the target domain. Furthermore, $n_{s}$ and $n_{t}$ stand for the sample size of the source domain and the target domain, respectively. As the source domain and target domain possess different marginal and conditional distributions, i.e., $p_{s} (\bm{X}_{s}) \neq p_{t} (\bm{X}_{t})$, the domain shift problem occurs. We assume that certain inherent similarities exist between different domains, which is valid as both domains share the same segmentation output space \citep{tsai2018learning,  mbatagan}.

As depicted in Fig.~\ref{fig2}, the proposed De-GLGAN is designed by capitalizing upon frequency decomposition and global-local context utilization. Specifically, multiscale HLFD modules are presented to align cross-domain representations by exploiting multiscale features generated by the encoder. Moreover, the global-local discriminator named GLDis further exploits the decomposed features to learn multiscale domain-invariant representations. In particular, the GLDis attempts to mimic the decoder to learn the domain-specific output space using the global-local representations.

\subsection{Encoder and Multiscale High/Low-Frequency Decomposition Modules}
\begin{figure}[t]
\centering
{\includegraphics[width=1\linewidth]{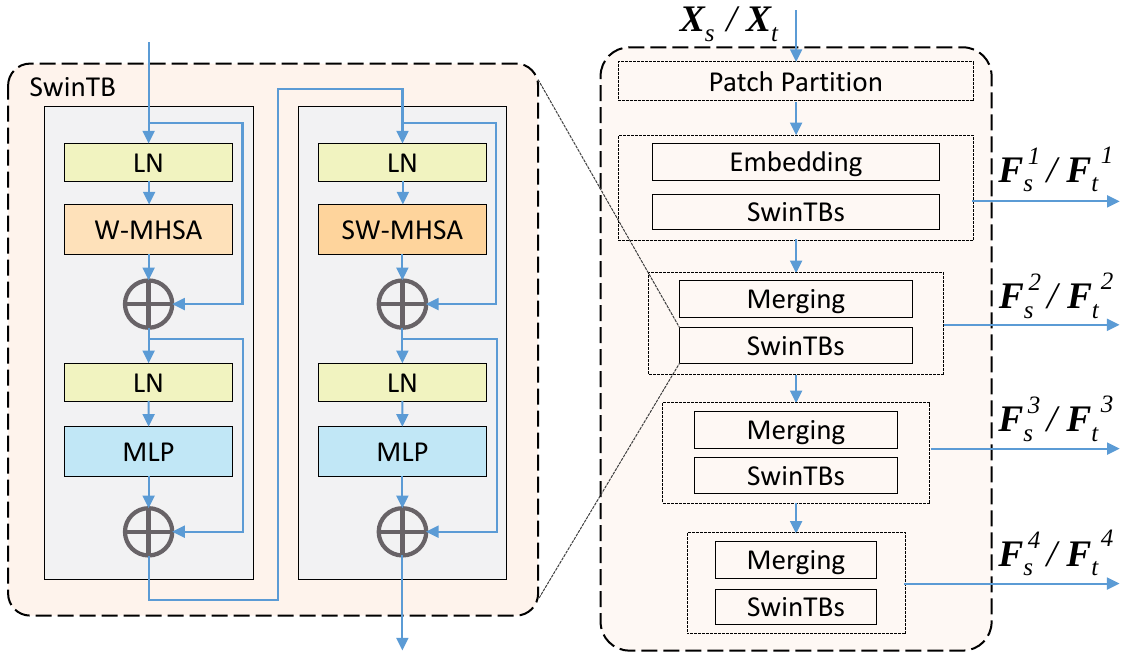}}
\caption{The detailed structure of the encoder and the SwinTB. The encoder extracts image features by stacking SwinTBs based on MHSA. By loading the pre-trained weights, it can effectively extract cross-domain multiscale image features.}
\label{fig3}
\end{figure}

As illustrated in Fig.~\ref{fig3}, the encoder comprises a Patch Partition layer and four successive stages. More specifically, the first stage contains one embedding layer followed by the SwinTBs, whereas the following three stages are equipped with one merging layer followed by each SwinTB. We denote by $\bm{X}_{s}\in \mathbb{R}^{{H\times W \times Z}}$ and $\bm{X}_{t}\in \mathbb{R}^{{H\times W \times Z}}$ the training images from the source and target domains, respectively. Furthermore, $Z$ stands for the number of image channels, whereas $H$ and $W$ represent the height and width of the image, respectively. $\bm{X}_{s}$ and $\bm{X}_{t}$ are first split into non-overlapping patches of size $4 \times 4$ by the Patch Partition layer. After that, the first-stage Embedding layer is applied to project these patches onto the encoding space before the SwinTBs gather image information on these patch tokens. As shown in Fig.~\ref{fig3}, the SwinTB consists of four LN layers, two MLPs, one W-MHSA and one shifted window-based multi-head self-attention (SW-MHSA) \citep{vaswani2017attention, liu2021swin}. After the first stage, three more successive stages are applied to generate the multiscale features denoted as $\bm{F}_{s}^{\emph {1-4}}$ and $\bm{F}_{t}^{\emph {1-4}}$ both of size $d_{i} \times \frac{H}{2^{i + 1}}\times \frac{W}{2^{i + 1}}$ where $d_{i}$ is the encoding dimension with $i\in\left\{1,2,3,4\right\}$ being the stage index. After that, these multiscale features are then fed into the corresponding HLFD module for frequency decomposition. Notably, the images from the source and the target domain are fed into the generator successively, that is, $\bm{F}_{s}^{\emph {1-4}}$ and $\bm{F}_{t}^{\emph {1-4}}$ are generated and collected, respectively.

\begin{figure}[t]
\centering
{\includegraphics[width=\linewidth]{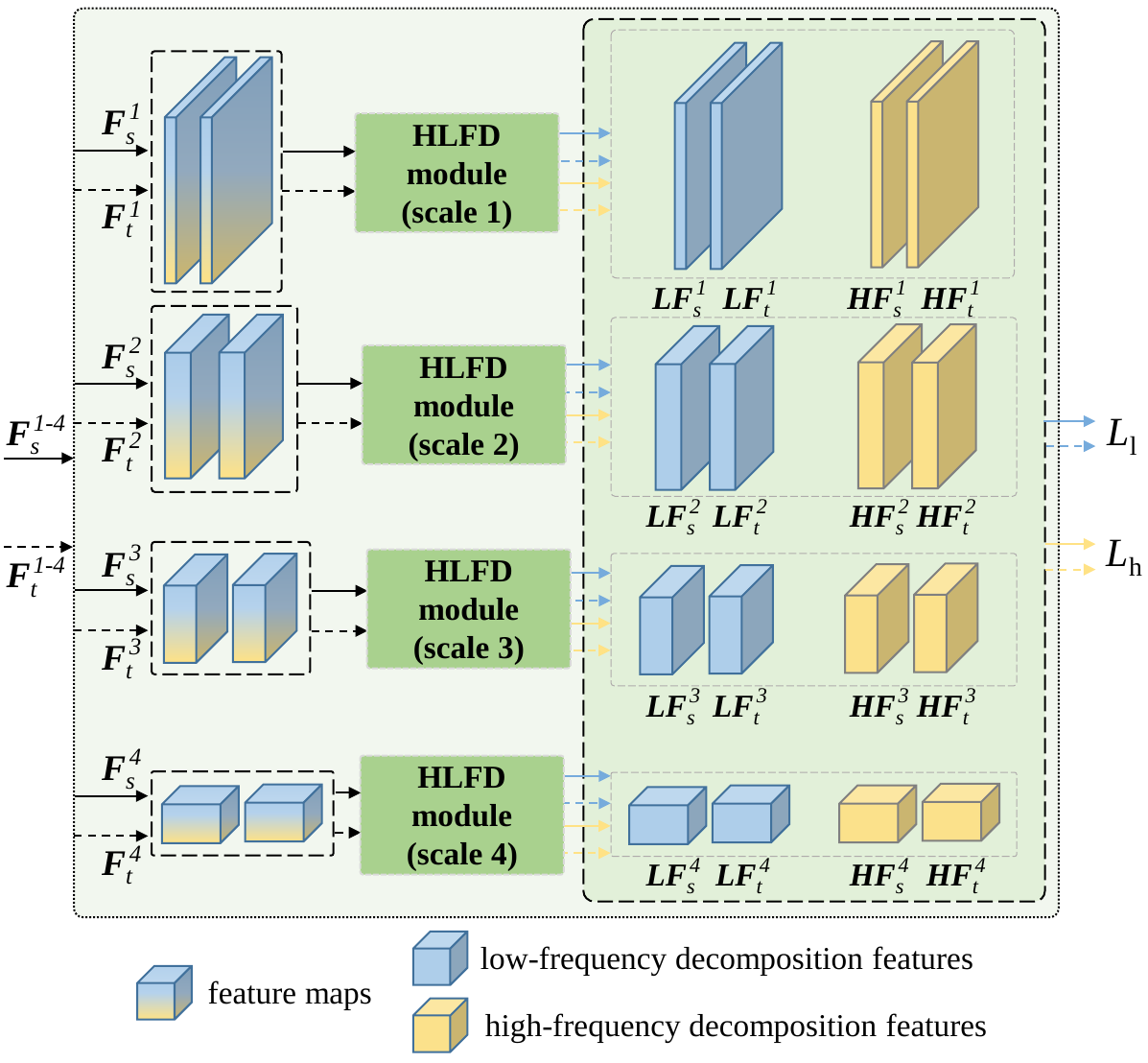}}
\caption{The structure of the proposed multiscale HLFDs. The multiscale features from the source and target domains are decomposed by the corresponding scale HLFD into low-frequency and high-frequency decomposition features. The superscript $*(=\emph{1,2,3,4})$ represents the index of multiscale features.}
\label{fig4}
\end{figure}

HLFD modules are employed to decompose the multiscale feature maps $\bm{F}_{s}^{\emph {1-4}}$ and $\bm{F}_{t}^{\emph {1-4}}$, as illustrated in Fig.~\ref{fig4}. The feature map at each scale, including the features from the source domain and the target domain, will be decomposed into the high or low-frequency feature maps by the corresponding scale HLFD. After that, the final outputs of the HLFD modules denoted as $\bm{LF}_{s}^{\emph {*}}/\bm{LF}_{t}^{\emph {*}}$ and $\bm{HF}_{s}^{\emph {*}}/\bm{HF}_{t}^{\emph {*}}$ are used to compute the frequency alignment loss $L_{\rm{f}}$, which will be elaborated in Sec.~\ref{Loss}.

\subsection{Global-Local Decoder}
\begin{figure}[htp]
\centering
{\includegraphics[width=1.0\linewidth]{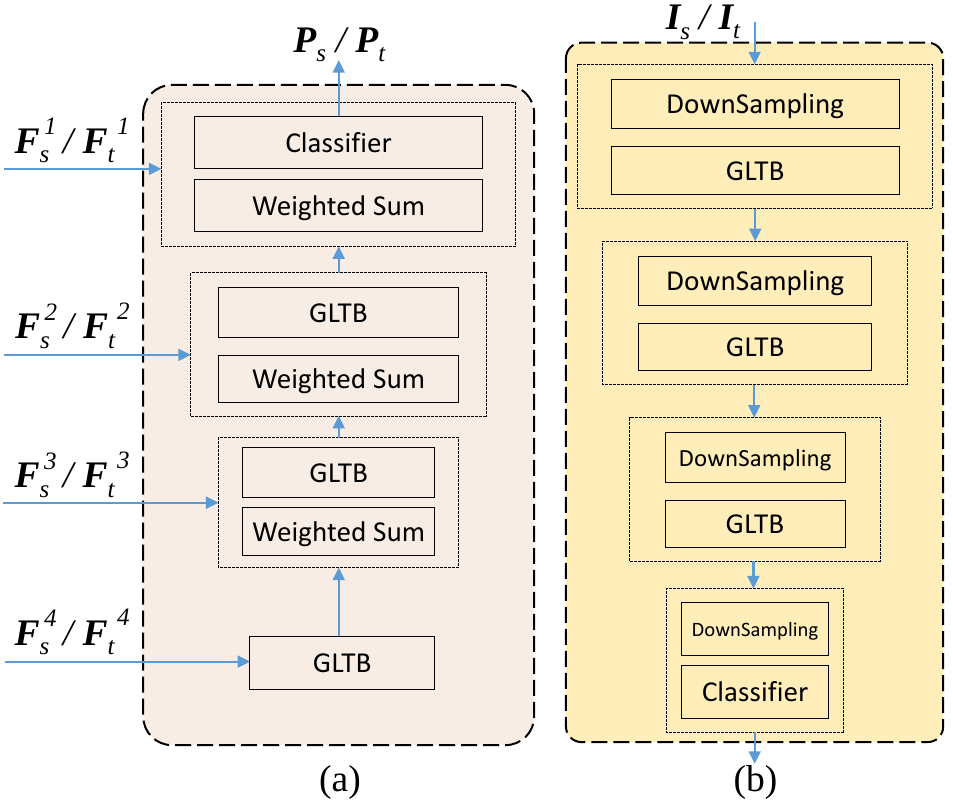}}
\caption{The detailed structures of (a) the decoder, and (b) the GLDis. They are similar in structure but different in function. The former is used for image decoding, while the latter is used for cross-domain information judgment.}
\label{fig4_2}
\end{figure}

The decoder is developed based on GLTB \citep{wang2022unetformer} and designed to exploit the decomposed features by gradually extracting global and local information, as shown in Fig.~\ref{fig4_2}(a). Specifically, the high-level feature maps derived from the encoder are first fed into an individual GLTB whose output is processed by two stages of weighted sum operation and GLTB. In particular, GLTB is able to capture the global context and local details simultaneously. After that, the weighted sum operation is employed to extract domain-invariant features by adaptively fusing the residual features and those from the previous GLTB. After the first three stages of the GLTB-based decoding, a weighted sum operation is used to fuse the residual connection from the first stage of the encoder and the deep global-local feature derived from previous GLTB before a classifier produces the final segmentation maps denoted by $\bm{P}_{s}$ and $\bm{P}_{t}$ of size $\mathbb{R}^{H\times W \times C}$ for the source and target domains, respectively, where $C$ stands for the number of ground object categories.

Finally, the supervised segmentation loss $L_{\rm{seg}}$ and the adversarial loss $L_{\rm{adv}}$ are also derived in the generator optimization stage. More specifically, $L_{\rm{seg}}$ is computed based on the source labels while $L_{\rm{adv}}$ is calculated using the adversarial strategy. More details about the loss functions will be provided in Sec.~\ref{Loss}.

\subsection{Global-Local Discriminator}\label{dis}
In the proposed discriminator, GLDis, we employ GLTB to exploit the semanftic output space produced by the generator. As shown in Fig.~\ref{fig4_2}, the GLDis and the decoder are similar in structure but different in function. Specifically, it first down-samples the weighted self-information (WSI) distribution $\bm{I}_{s}$ and $\bm{I}_{t}$ derived from $\bm{P}_{s}$ and $\bm{P}_{t}$ before feeding the down-sampled data into a GLTB. Given the output segmentation maps of the {\em source} domain $\bm{P}_{s}$ and the {\em target} domain $\bm{P}_{t}$, the corresponding self-information is defined as $-\mathrm{log} \bm{P}_{s}$ and $-\mathrm{log} \bm{P}_{t}$, respectively. Consequently, the WSI maps $\bm{I}_{s}$ and $\bm{I}_{t}$ are calculated as follows:
\begin{eqnarray}
    \bm{I}_{s}^{(h, w)} = -\bm{P}_{s}^{(h, w)}\cdot\mathrm{log} (\bm{P}_{s}^{(h, w)}),\label{eq:self-infs}\\
		\bm{I}_{t}^{(h, w)} = -\bm{P}_{t}^{(h, w)}\cdot\mathrm{log} (\bm{P}_{t}^{(h, w)}),\label{eq:self-inft}
\end{eqnarray}
where the dimensions of $\bm{P}_{s}$, $\bm{P}_{t}$, $\bm{I}_{s}$ and $\bm{I}_{t}$ are $\mathbb{R}^{H\times W \times C}$, and $(h, w)$ represents the index position of one pixel. These information maps are regarded as the disentanglement of the Shannon Entropy. After the first GLTB, the resulting data is further processed by two successive stages of down-sampling, followed by the GLTB, which is designed to gradually gather the global and local information in the segmentation map-generated features. Finally, the resulting data is down-sampled and processed by a classifier, which predicts the domain to which the input segmentation maps belong. Despite its appearance similar to the decoder structure, GLDis is tailored to learn domain-specific representations using an adversarial strategy. More specifically, the decoder upsamples the abstract semantic features, whereas GLDis down-samples the structured output space to identify the domain to which each input belongs. The final output of GLDis is used to compute a cross-entropy loss denoted by $L_{\rm{d}}$ in the discriminator training stage. More details about the loss function will be presented in the next section.

\subsection{Loss Functions}\label{Loss}
The proposed De-GLGAN capitalizes on four loss functions, namely the segmentation loss $L_{\rm{seg}}$, the typical adversarial loss $L_{\rm{adv}}$, the frequency alignment loss $L_{\rm{f}}$, and the cross-entropy loss $L_{\rm{d}}$. These loss functions are defined as follows.

\subsubsection{Generator} $L_{\rm{seg}}$, $L_{\rm{f}}$ and $L_{\rm{adv}}$ are proposed to optimize the generator. In particular, $L_{\rm{seg}}$ is computed directly using images from the {\em source} domain and their corresponding labels. Mathematically, $L_{\rm{seg}}$ takes the following form:
    \begin{eqnarray}
    L_{\rm{seg}} (\bm{P}_{s})=-\lambda^{seg}\sum_{h, w} \sum_{c \in {\mathcal C}}\bm{Y}_{s}^{(h, w, c)}\left[\mathrm{log} (\bm{P}_{s}^{(h, w, c)})\right],\label{eq:seg_loss}
    \end{eqnarray}
where ${\mathcal C}$ is the category set whose cardinality is the number of categories $C$, $\bm{Y}_{s}$ of size $\mathbb{R}^{H\times W \times C}$ denotes the labels from the source domain, and $\lambda^{seg}$ is a scalar, representing the weighting coefficient for the segmentation loss.
$L_{\rm{f}}$ is computed to align decomposed cross-domain features generated by multiscale HLFDs as follows:
\begin{equation} 
		\begin{split}
		&  L_{\rm{f}}\left(\bm{LF}_{s}^{\emph {1-4}}, \bm{LF}_{t}^{\emph {1-4}}, \bm{HF}_{s}^{\emph {1-4}}, \bm{HF}_{t}^{\emph {1-4}}\right)\\
		& = L_{\rm{l}}\left(\bm{LF}_{s}^{\emph {1-4}}, \bm{LF}_{t}^{\emph {1-4}}\right) + L_{\rm{h}}\left(\bm{HF}_{s}^{\emph {1-4}}, \bm{HF}_{t}^{\emph {1-4}}\right)\\
		& = \lambda^{f}\cdot\displaystyle\sum_{i=1}^{\emph {4}}\displaystyle\sum_{d=1}^{d_{i}}\frac{\left(\mathbf{P} (\bm{LF}_{s}^{i,d})-\mathbf{P} (\bm{LF}_{t}^{i,d})\right)}{d_{i}}\\
		& + \lambda^{f}\cdot\displaystyle\sum_{i=1}^{\emph {4}}\displaystyle\sum_{d=1}^{d_{i}}\frac{\left(\mathbf{P} (\bm{HF}_{s}^{i,d})-\mathbf{P} (\bm{HF}_{t}^{i,d})\right)}{d_{i}},\label{eq:fre}
		\end{split}
\end{equation}
where the sizes of $\bm{LF}_{s}^{i}$, $\bm{LF}_{t}^{i}$, $\bm{HF}_{s}^{i}$ and $\bm{HF}_{t}^{i}$ are $d_{i} \times \frac{H}{2^{i + 1}}\times \frac{W}{2^{i + 1}}$, $d_{i}$ is the encoding dimension with $i\in\left\{1,2,3,4\right\}$ being the stage index, $\lambda^{f}$ is a scalar, representing the weighting coefficient, and $\mathbf{P}$ stands for the average pooling, which gathers spatial information for each channel $d$. The frequency alignment loss function is designed to efficiently align the feature maps across the source and target domains.

To align the structured output space, the WSI distribution is adopted in the proposed training process to calculate the adversarial loss as follows:
\begin{eqnarray}
L_{\rm{adv}}\left(\bm{I}_{t}\right)=-\lambda^{adv}\sum_{h, w}\left[\mathrm{log}\left(\mathbf{D} (\bm{I}_{t})^{(h, w, 0)}\right)\right],\label{eq:adv_loss}
\end{eqnarray}
where $\lambda^{adv}$ is a scalar, representing the weighting coefficient for the adversarial loss while $\mathbf{D}$ denotes the GLDis. Note that the artificial labels $0$ and $1$ are assigned to samples drawn from the {\em source} domain and the {\em target} domain, respectively. In Eq.~\eqref{eq:adv_loss}, the label $0$ in the superscript $^{(h, w, 0)}$ is set to intentionally mislead GLDis. Therefore, the discriminator is motivated to bridge the prediction gap between the source and target domains.

Finally, the following overall objective function is proposed to optimize the generator:
\begin{equation}
  	L_{\rm{G}} = L_{\rm{seg}}+L_{\rm{f}}+L_{\rm{adv}}.\label{eq:gen}
\end{equation}

\subsubsection{GLDis} The discriminator GLDis is optimized with the following cross-entropy loss $L_{\rm{d}}$ to identify to which domain a given segmentation map belongs.  
\begin{equation}
    L_{\rm{d}} (\bm{I}_{s}, \bm{I}_{t})=-\sum_{h, w}\left[\mathrm{log}(\mathbf{D} (\bm{I}_{s})^{(h, w,0)})+\mathrm{log}(\mathbf{D} (\bm{I}_{t})^{(h, w,1)})\right].\label{eq:ce_loss}
\end{equation}
	
\subsubsection{Overall Loss Function} Finally, the overall loss function $L$ proposed for De-GLGAN is given as follows:
\begin{equation}
	\begin{aligned}
		\mathop{\max}_{\mathbf{D}}\ \mathop{\min}\limits_{\mathbf{G}} L(\bm{X}_{s}, \bm{Y}_{s}, \bm{X}_{t}),\label{eq0}
	\end{aligned}
\end{equation}
where $\bm{X}_{s}$ and $\bm{X}_{t}$ are with the size of $\mathbb{R}^{{H\times W \times Z}}$, and $\mathbf{D}$ and $\mathbf{G}$ denote GLDis and the generator, respectively. Eq.~\eqref{eq0} is designed to minimize the segmentation loss in the generator for the source images while maximizing the output space similarity of the source and target domains. Notably, the generator and the discriminator are trained in a classical GAN-based manner, whereas the generator and GLDis are trained alternatively. More specifically, the generator is first trained using Eq.~(\ref{eq:gen}) with the parameters of GLDis being held constant before GLDis is updated with Eq.~(\ref{eq:ce_loss}). The overall optimization process of De-GLGAN is presented in Algorithm 1 for better understanding. Finally, the well-trained generator $\hat{\mathbf{G}}$ is used for testing in the target domain.
\begin{algorithm}[!h]
    \caption{Optimization Process of De-GLGAN}
    \label{alg:opt}
    \renewcommand{\algorithmicrequire}{\textbf{Input:}}
    \renewcommand{\algorithmicensure}{\textbf{Output:}}
    \begin{algorithmic}[1]
        \REQUIRE Labeled source dataset $D_{s}$; Unlabeled target dataset $D_{t}$; Initialized  generator $\mathbf{G}$, multiscale HLFDs and discriminator $\mathbf{D}$; Weighting coefficients $\lambda^{seg}$, $\lambda^{f}$ and $\lambda^{adv}$. 
        \ENSURE Well-trained generator $\hat{\mathbf{G}}$.
        \WHILE{in total epochs}
            \STATE Sample a mini-batch $\bm{X}_{s}$, $\bm{Y}_{s}$ and $\bm{X}_{t}$;
						\STATE Input $\bm{X}_{s}$ into $\mathbf{G}$ to obtain $\bm{F}_{s}^{\emph {1-4}}$ and $\bm{P}_{s}$;
						\STATE Input $\bm{X}_{t}$ into $\mathbf{G}$ to obtain $\bm{F}_{t}^{\emph {1-4}}$ and $\bm{P}_{t}$;
						\STATE Input $\bm{F}_{s}^{\emph {1-4}}$ and $\bm{F}_{t}^{\emph {1-4}}$ into multiscale HLFDs to obatin $\bm{LF}_{s}^{\emph {*}}/\bm{LF}_{t}^{\emph {*}}$ and $\bm{HF}_{s}^{\emph {*}}/\bm{HF}_{t}^{\emph {*}}$;
						\STATE Calculate $\bm{I}_{s}$ and $\bm{I}_{t}$ from $\bm{P}_{s}$ and $\bm{P}_{t}$ based on Eq.~\eqref{eq:self-infs} and Eq.~\eqref{eq:self-inft};
						\STATE Calculate $L_{\rm{seg}}$ based on Eq.~\eqref{eq:seg_loss};
						\STATE Calculate $L_{\rm{f}}$ based on Eq.~\eqref{eq:fre};
						\STATE Fix $\mathbf{D}$ and input $\bm{I}_{t}$ into $\mathbf{D}$ to calculate $L_{\rm{adv}}$ based on Eq.~\eqref{eq:adv_loss};
						\STATE Update $\mathbf{G}$ based on Eq.~\eqref{eq:gen};
						\STATE Unfix $\mathbf{D}$ and input $\bm{I}_{s}$ and $\bm{I}_{t}$ into $\mathbf{D}$ to calculate $L_{\rm{d}}$ based on Eq.~\eqref{eq:ce_loss};
						\STATE Update $\mathbf{D}$ based on Eq.~\eqref{eq:ce_loss}.
        \ENDWHILE
        \RETURN $\hat{\mathbf{G}}$ = $\mathbf{G}$.
    \end{algorithmic}
\end{algorithm}
\section{Experiments And Discussions}\label{sec:experiment}
\subsection{Datasets Description}
\subsubsection{ISPRS Potsdam and Vaihingen} This dataset \citep{gerke2014use} involves fine-resolution images obtained from two cities. 
The images in both datasets were obtained from aerial photography but of different Ground Sampling Distance (GSD). More specifically, the images of Potsdam have clearer ground objects with a GSD of $5$~cm as compared to a GSD of $9$~cm for Vaihingen. Furthermore, images in Potsdam possess richer spectrum information from four frequency bands, namely, InfraRed, Red, Green and Blue (IRRGB), whereas images in Vaihingen only three bands, namely, InfraRed, Red and Green channels (IRRG). All images in ISPRS Potsdam and Vaihingen are provided with semantic labels, including six ground object classes, namely, {\em Building (Bui.)}, {\em Tree (Tre.)}, {\em Low Vegetation (Low.)}, {\em Car}, {\em Impervious Surface (Imp.)} and {\em Clutter/Background}.

\textbf{Potsdam}: There are $24$ very fine-resolution true orthophotos of size $6000 \times 6000$ in Potsdam dataset. In the experiments, we divided these orthophotos into a training set of $18$ images and a test set of $6$ images.

\textbf{Vaihingen}: The Vaihingen dataset contains $16$ very fine-resolution true orthophotos of size $2500 \times 2000$ pixels. In our experiments, these orthophotos are divided into a training set of $12$ images and a test set of $4$ images.

\begin{figure}[t]
	\centering
	{\includegraphics[width=0.95\linewidth]{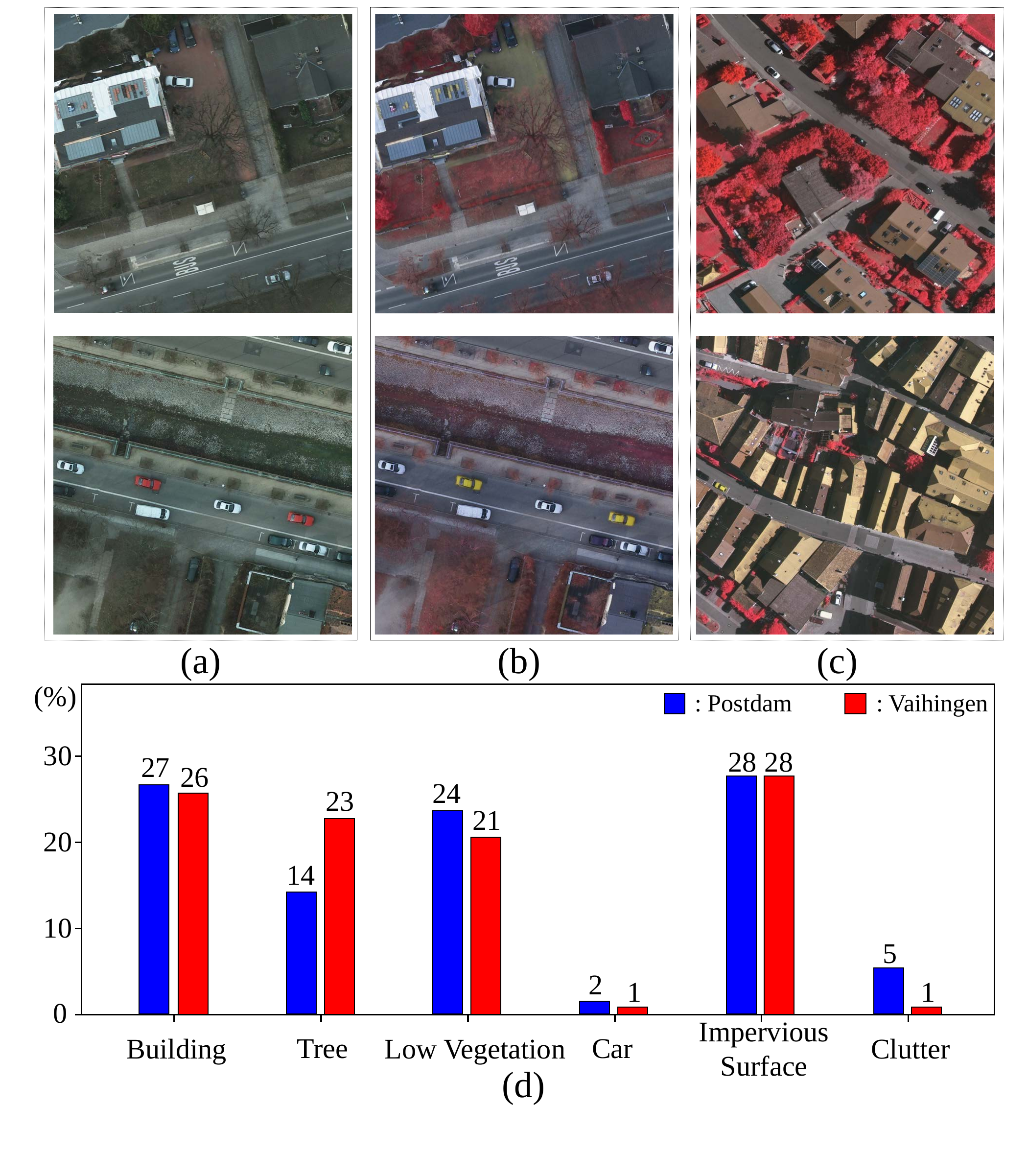}}
	\caption{Image samples of size $1024 \times 1024$ from (a) P-RGB, (b) P-IRRG and (c) V-IRRG, respectively. 
Despite some visual similarities between these two datasets, they exhibit large discrepancies in the distribution of land objects. (d) Class distribution statistics for the Potsdam and Vaihingen.}
	\label{fig5_0}
\end{figure}

Fig.~\ref{fig5_0}(a-c) illustrates two groups of images sampled from the Potsdam and Vaihingen datasets. Visual inspection of Fig.~\ref{fig5_0}(a-c) confirms large discrepancies across the two datasets. However, since the scene categories of Potsdam and Vaihingen are similar, the category distributions are close as shown in Fig.~\ref{fig5_0}(d). Therefore, the domain shift in UDA tasks constructed on Potsdam and Vaihingen mainly comes from geographical location and image mode. In the sequel, we divide the Potsdam dataset into two subsets, namely, ``P-IRRG" and ``P-RGB", taking into account the data size and image channels as compared to those of Vaihingen. P-IRRG contains infrared, red, green bands while P-RGB contains red, green and blue bands. The Vaihingen dataset forms only one dataset denoted as ``V-IRRG" including infrared, red, green bands. By treating these three subsets as the source and target domains, we design the first three UDA tasks with increasing levels of difficulty as listed in Table~\ref{tab:tasks}. 

It is worth noting that these three tasks exhibit different levels of domain shifts. For instance, Task $1$, i.e., P-IRRG to V-IRRG, suffers from the domain shift caused by large discrepancies in the spatial resolution, the ground object style and the illumination environment, though the images from both domains possess the same spectrum information. In contrast, the domain shift problem in Task $2$, i.e., P-RGB to V-IRRG, is caused by the frequency band difference in addition to all discrepancies in Task $1$. Finally, Task $3$, i.e., V-IRRG to P-RGB, encounters the same domain shift problem as Task $2$ with a greater challenge as the size of the Vaihingen dataset in the source domain is much smaller than that of the Potsdam dataset in the target domain. In summary, these three UDA tasks were designed to explore the effectiveness of the UDA method under consideration in cross-location, in-image mode, and cross-image mode UDA tasks.

\begin{figure}[t]
	\centering
	{\includegraphics[width=0.95\linewidth]{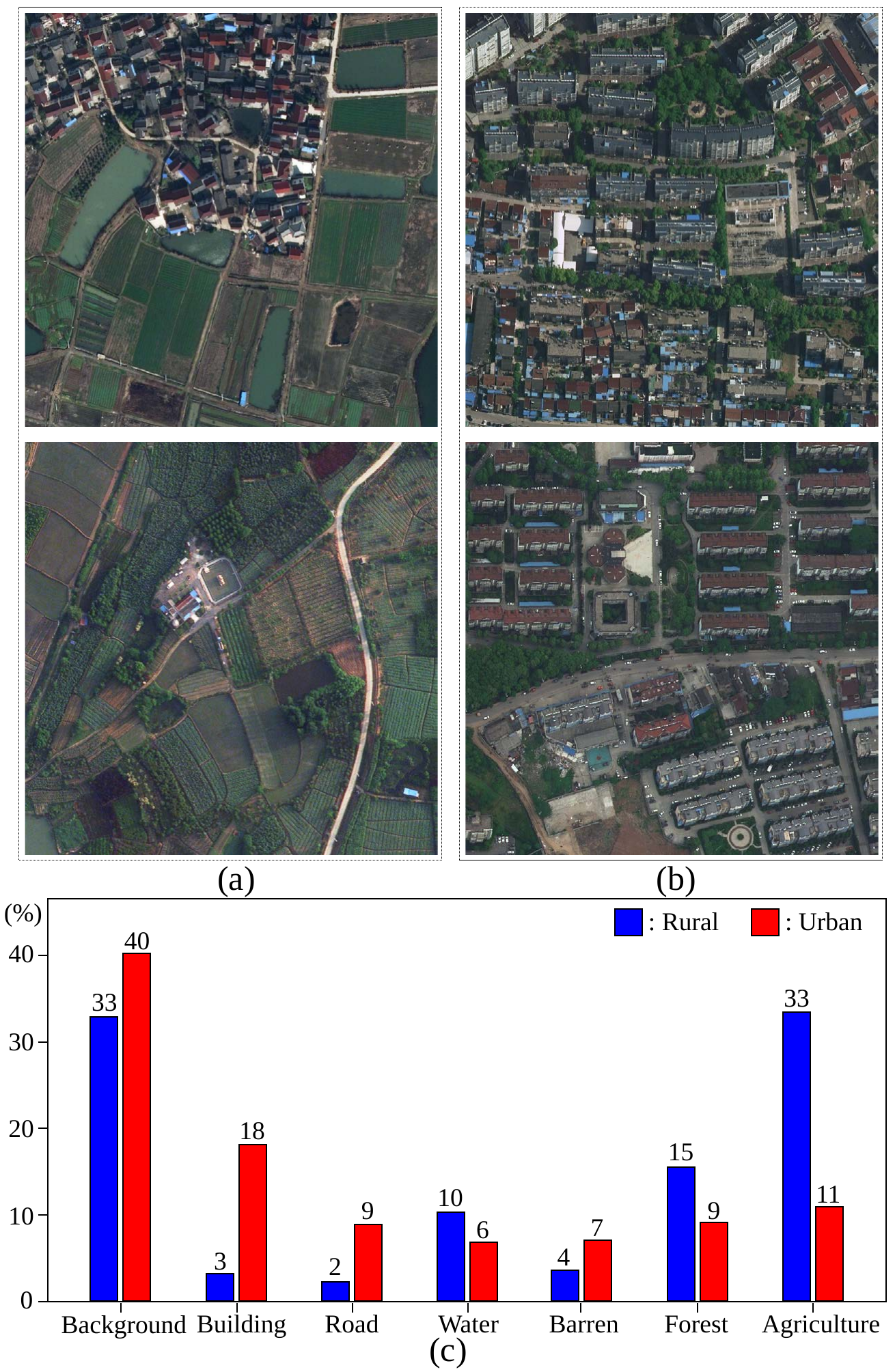}}
	\caption{Image samples of size $1024 \times 1024$ from (a) LoveDA Rural and (b) LoveDA Urban, respectively. (c) Class distribution statistics for the Rural and Urban scenes. The main difference between these two scenes are the large intra-class variance and the inconsistent class distributions.}
	\label{fig5_1}
\end{figure}

\subsubsection{LoveDA Rural and Urban}
The LoveDA dataset contains two scenes, namely Rural and Urban. The Rural and Urban comprise $2358$ and $1833$ high-resolution optical remote sensing images, respectively. All images are with the size of $1024\times 1024$ pixels. The images provide three channels, namely Red, Green, and Blue (RGB), with a ground sampling distance of $30$ cm. The dataset encompasses seven landcover categories, including {\em Background (Bac.)}, {\em Building (Bui.)}, {\em Road}, {\em Water (Wat.)}, {\em Barren (Bar.)}, {\em Forest (For.)}, and {\em Agriculture (Agr.)} \cite{wang2021loveda}. These images were collected from three cities in China (Nanjing, Changzhou, and Wuhan).

\textbf{Rural}: 
The $2358$ images are divided into two parts, with $1366$ images for training and $992$ images for testing. Specifically, the training set contains images indexed from $0$ to $1365$, while the test set spans images from $2522$ to $3513$.

\textbf{Urban}: 
We divided the $1833$ images into $1156$ images for training and $677$ images for testing. The former contains images indexed from $1366$ to $2521$, while the latter spans images from $3514$ to $4190$.

Fig.~\ref{fig5_1}(a) and (b) present two groups of images sampled from the LoveDA Rural and Urban scenes. The discrepancy between the two scenes primarily arises from variations in ground object categories caused by geographical location. To be specific, there is intra-class variance across scene ground objects, such as shape, layout, and scale. Additionally, the class distributions are inconsistent between rural and urban scenes as shown in Fig.~\ref{fig5_1}(c). These differences bring great challenges to the learning of transferable features. Following the common practice reported in the literature \citep{wang2021loveda, ma2023domain}, we conducted the last two UDA tasks as listed in Table~\ref{tab:tasks}.
During the training stage, the size of the sliding window was set to $256 \times 256$ to collect the training samples dynamically. Therefore, we can obtain $1280$ Vaihaigen images, $13824$ Potsdam images, $37728$ Rural images and $29328$ Urban images.

\begin{table}[h]
	\begin{center}
		\caption{UDA Tasks}
		\label{tab:tasks}
		\begin{tabular}{c|c|c} 
			\hline
			&\textbf{Source Domain} & \textbf{Target Domain} \\\hline
			Task 1& P-IRRG & V-IRRG\\
			Task 2& P-RGB  & V-IRRG\\
			Task 3& V-IRRG & P-RGB\\
			\hline
			Task 4& Rural  & Urban\\
			Task 5& Urban  & Rural\\
			\hline
		\end{tabular}
	\end{center}
\end{table}

\begin{table*}[t]
	\centering
	\caption{Eleven methods were evaluated in our following experiments, including two non-UDA and nine UDA methods.}
	\begin{tabular}{|c|l|l|p{12cm}|}\hline
		Index&Method & Reference & Description  \\\hline
		1&Source-only & \citep{jha2019resunet++} & Images and labels from the source domain are utilized to train the segmentation network before it is tested directly on the target domain.\\\hline
		2&DANN & \citep{ganin2016domain} & The first UDA method that applies adversarial learning to solve the domain shift problem on the image classification task.\\\hline
		3&AdasegNet &  \citep{tsai2018learning} & This UDA method performs multi-level adversarial learning with two discriminators while assuming that the source and target domains share a similar output space.\\\hline
		4&Advent & \citep{vu2019advent} & This UDA method aligns the entropy maps of the output derived by the generator for both domains.\\\hline
		5&GANAI& \citep{benjdira2019unsupervised} & The first UDA method designed for semantic segmentation of remote sensing images based on the GAN.\\\hline
		6&TriADA& \citep{yan2019triplet}& This UDA method employs a triplet branch for remote sensing images, considering both domains simultaneously to learn a domain-invariant classifier using a domain similarity discriminator.\\\hline
		7&CCAGAN& \citep{chen2022unsupervised}&  This UDA method, designed for remote sensing images, develops a category-certainty attention module focusing on the misaligned regions in the category level.\\\hline
		8&MBATA-GAN& \citep{mbatagan}& This UDA method develops a mutually boosted attention-based transformer to augment the high-level features and a feature discriminator to learn the transferable features of remote sensing images.\\\hline
		9&MASNet& \citep{zhu2023unsupervised}& This UDA method stores domain-invariant prototypical representations by employing a feature memory module, bridging the domain distribution discrepancy of two remote sensing domains. \\\hline
		10&DSSFNet& \citep{chen2024unsupervised}& This UDA method decouples ground object features to learn domain-invariant semantic representations and domain-specific style information for remote sensing images. \\\hline
		11&Supervised& \citep{jha2019resunet++}& Images and labels from the target domain are used to train and test the segmentation network, which stands for the optimal performance that any UDA method can achieve. \\\hline
	\end{tabular}\label{tab:methods}
\end{table*}

All models in the experiments were implemented with the PyTorch framework on a single NVIDIA GeForce RTX 3090 GPU with $24$-GB RAM. Furthermore, the generator was optimized with the Stochastic Gradient Descent (SGD) optimizer and the Nesterov acceleration with a momentum of $0.9$ and a weight decay of $5\times10^{-4}$. The initial learning rate was set to $2.5\times10^{-4}$, decreasing with a polynomial decay of power $0.9$. For GLDis, the Adam optimizer was employed with the learning rate of $10^{-4}$ with a polynomial decay of power $0.9$. The momentum of Adam is set to $\left[0.9, 0.99\right]$. In addition, the multi-level adversarial learning scheme was also adopted with parameters set to their default values derived from AdasegNet \citep{tsai2018learning} in which $\lambda^{seg}$ and $\lambda^{adv}$ were set to $1.0$ and $0.001$, respectively. Finally, $\lambda^{f}$ was set to $0.01$ according to the parameter analysis that will be presented in Sec.~\ref{sec:para}.
\subsection{Evaluation Metrics}
In this work, we adopt three metrics to quantitatively assess the segmentation performance of the cross-domain remote sensing images for all UDA methods, namely, the Overall Accuracy (OA), the F1 score (F1) and the Intersection over Union (IoU), which are defined as follows:
\begin{eqnarray}
&OA = \frac{TP+TN}{TP+TN+FP+FN},\\
&Precision=\frac{TP}{TP+FP},\\
&Recall=\frac{TP}{TP+FN},\\
&F1 = 2 \times \frac{Precision \times Recall}{Precision + Recall},\\
&IoU = \frac{TP}{TP+FP+FN},
\end{eqnarray}
where $TP$, $FP$, $TN$ and $FN$ are true positives, false positives, true negatives and false negatives, respectively. Furthermore, the mean F1 score (mF1) and the mean IoU (mIoU) for the main five classes apart from {\em Clutter/Background} are calculated to measure the average performance of the UDA methods following the common practice reported in the literature \citep{yan2019triplet, mbatagan}.

In addition, the computational complexity of the proposed GLGAN and De-GLGAN is evaluated using the following metrics. First, the giga floating point operation counts (GFLOPs) are employed to evaluate the model complexity. Second, the number of model parameters and the memory footprint are utilized to evaluate the memory requirement. Finally, the frames per second (FPS) is used to evaluate the execution speed. For a computationally efficient method, its first three metrics are usually small, while its FPS value should be large.
Each method undergoes training for $50$ epochs, with each epoch consisting of $1000$ batches. During training, the performance of each method is evaluated every 100 batches and recorded. The best-performing model for each method is then selected for subsequent testing based on its highest segmentation performance.

\subsection{Experimental settings}
In our work, the basic unit of the encoding block in the generator is adopted from the Swin Transformer \citep{liu2021swin} pretrained on ImageNet. In the experiments, two versions of Swin Transformers, namely Swin-Base and Swin-Large, were applied for image feature extraction. Both Swin-Base and Swin-Large have four stages with the stack of $\left\{2,2,18,2\right\}$ SwinTBs with the embedding dimensions of Swin-Base and Swin-Large being $\left\{128,256,512,1024\right\}$ and $\left\{196,384,768,1536\right\}$, respectively. Furthermore, the number of the heads in Swin-Base and Swin-Large are $\left\{4,8,16,32\right\}$ and $\left\{6,12,24,48\right\}$, respectively. In addition, the window size for W-MHSA and SW-MHSA is $16\times16$. The dimension of both the decoder and GLDis is set to $256$, whereas the window size of the global-local attention is $8\times8$. 

\subsection{Benchmarking}
To confirm the effectiveness of the proposed methods, some representative existing UDA methods were selected for quantitative comparison. Table~\ref{tab:methods} summarizes the nine methods evaluated in our following experiments. Notably, in addition to the first three representative methods in the field of computer vision, the other six methods are specifically designed for remote sensing images. Furthermore, we also performed comparison experiments on two non-DA methods for the first three UDA tasks on the ISPRS Potsdam and Vaihingen dataset, namely a deep model without domain adaptation (labeled as ``Source-only") and a supervised deep model (labeled as ``Supervised") trained with labeled data from the target domain. These two non-DA methods stand for the two extreme cases, with the ``Supervised" model representing the optimal performance that any UDA method targets and the ``Source-only" model standing for the performance lower bound. Furthermore, the classical ResUNet++ \citep{jha2019resunet++} is selected for these two non-DA methods by taking into account the balance between segmentation performance and computational complexity. 
For the aforementioned methods, the same network architectures, training parameters, and evaluation methods were applied to make a fair comparison.

\begin{table*}[t]\footnotesize
	\centering
	\caption{Task 1: Quantitative comparison of F1 (\%) on the adaptation of P-IRRG to V-IRRG.}
	\setlength{\tabcolsep}{5mm}{
		\begin{tabular}{m{2.7cm}<{\centering}|m{1.4cm}<{\centering}|m{0.4cm}<{\centering}m{0.4cm}<{\centering}m{0.4cm}<{\centering}m{0.4cm}<{\centering}m{0.4cm}<{\centering}|m{0.4cm}<{\centering}m{0.4cm}<{\centering}m{0.4cm}<{\centering}}
			\hline
			\textbf{Method}     & \textbf{Backbone} & \textbf{Bui.} & \textbf{Tre.}  & \textbf{Low.}  & \textbf{Car}   & \textbf{Imp.}  & \textbf{OA}    & \textbf{mF1}    & \textbf{mIoU}   \\
			\hline
			Non-DA (Source-Only)  &   ResUNet++  & 76.93 & 70.79 & 40.19 & 7.06  & 56.70 & 63.03 & 50.33 & 37.13  \\
   		\hline
			DANN         &  -   & 83.07 &	75.93 &	54.15 &	55.38 &	69.73 &	71.82 &	67.66 &	52.24   \\
			AdasegNet    &  ResNet101   & 87.87 & 79.44 & 56.73 & 57.59 & 76.10 & 76.29 & 71.55 & 57.14  \\
			Advent       &  ResNet101   & 90.85	& 78.78	& 58.31	& 55.07	& 79.53 &	77.63 &	72.51 &	58.68 \\
			GANAI        &  ResNet101   & 87.13 & 77.20 &	52.48 &	58.78 &	74.38 &	73.52 &	69.99 &	55.29 \\
			TriADA       &  ResNet101   & 89.34 &	79.06 &	47.50 &	65.72 &	81.11 &	77.56 &	72.55 &	58.88 \\
			CCAGAN       &  ResNet101   & 89.02 &	78.10 &	59.29 &	65.42 &	79.10 &	76.90 &	74.19 &	60.09 \\
			MBATA-GAN    &  ResNet101   & 90.98 & 79.51 & 63.41 & 66.38 & 82.80 & 80.75 & 76.77 & 63.50 \\
			MASNet       &  ResNet101   & 90.13 & 80.90 & 59.11 & 66.70 & 81.44 & 79.87 & 75.66 & 62.13 \\
			DSSFNet      &  ResNet101   & 92.05 & 80.76 & 60.44 & 66.29 & 82.68 & 80.67 & 76.45 & 63.27 \\
			\hline
			\multirow{2}{*}{GLGAN} & Swin-Base  &	91.95 &	80.31 &	67.71 & 65.89  & 84.18 & 81.97 & 78.01 & 65.04 \\
             & Swin-Large &	92.33 & \textbf{81.60} & 69.11 & 67.34  & 84.58 & 82.74 & 78.99 & 66.30 \\
			De-GLGAN  & Swin-Base  & \textbf{93.94} &	80.95 &	\textbf{70.06} & \textbf{70.40}  & \textbf{86.13} & \textbf{83.66} & \textbf{80.30} & \textbf{68.09} \\
			\hline
			Non-DA (Supervised)     & ResUNet++  & 94.84 & 90.15 & 78.25 & 80.16 & 90.47 & 89.66 & 86.77 & 77.20 \\
			\hline
	\end{tabular}}\label{tab:results1}
\end{table*}

\subsection{Results and Discussions}\label{sec:tasks} 
\begin{figure}[t]
	\centering
	{\includegraphics[width=0.95\linewidth]{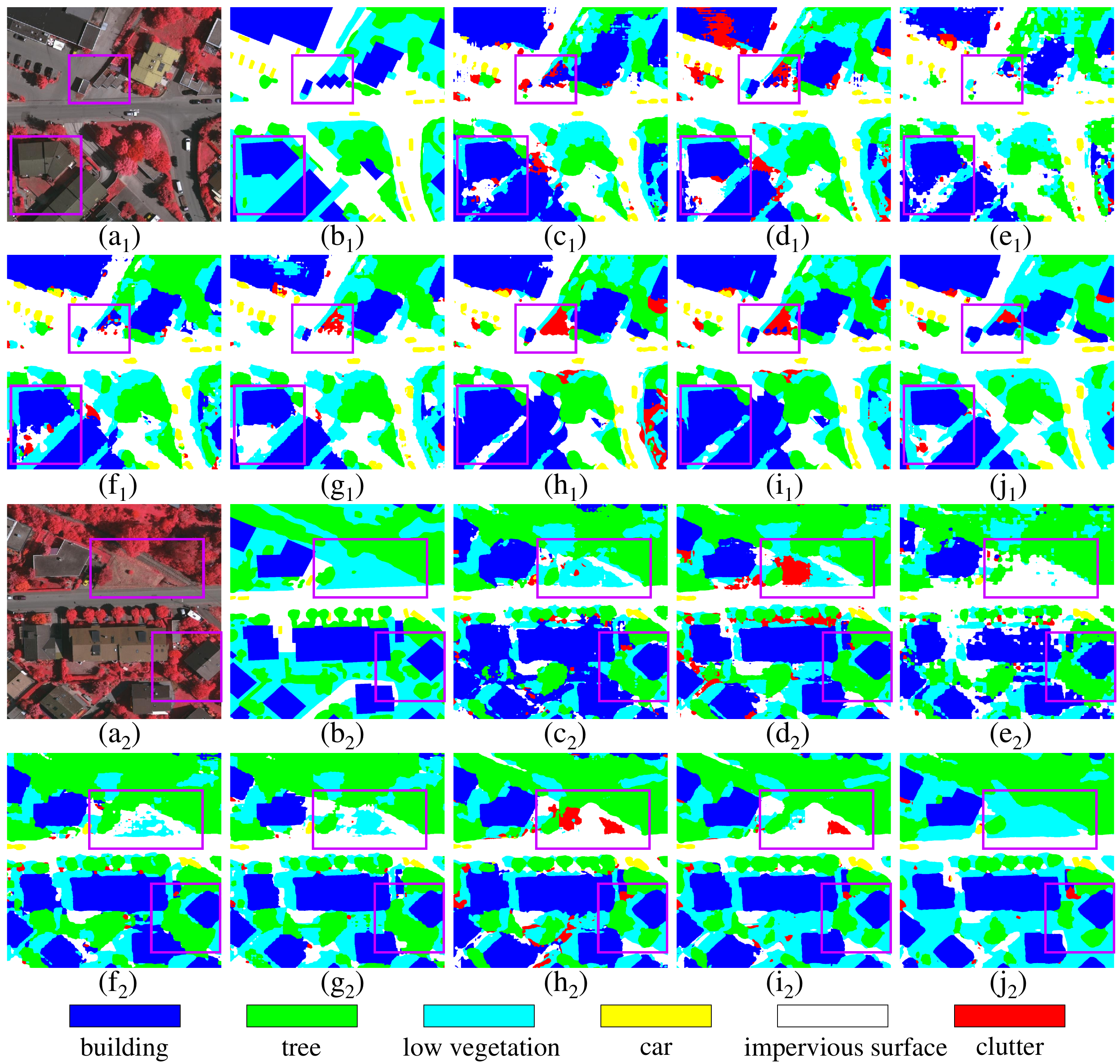}}
	\caption{Task 1: qualitative performance comparison of the adaptation P-IRRG to V-IRRG with images of size $1024 \times 1024$, where the two samples are distinguished by subscripts. (a) IRRG images, (b) Ground Truth, (c) AdasegNet, (d) Advent, (e) TriADA, (f) CCAGAN, (g) MBATA-GAN (h) MASNet, (i) DSSFNet and (j) De-GLGAN. Purple boxes are marked to highlight the differences.}
	\label{fig5}
\end{figure}

\noindent{\emph{\textbf{Task 1} (P-IRRG to V-IRRG):}}
The results for Task $1$ are summarized in Table~\ref{tab:results1}. Inspection of Table~\ref{tab:results1} suggests that the ``Source-Only" method showed the worst performance among all methods under consideration in terms of OA, mF1 and mIoU. Furthermore, all UDA methods demonstrated performance improvement of different degrees. For instance, the simplest DANN achieved improvements of $8.79\%$ in OA, $17.33\%$ in mF1 and $15.11\%$ in mIoU, which confirmed the necessity and effectiveness of the UDA approach. In particular, the baseline Advent showed improved OA, mF1, and mIoU of $77.63\%$, $72.51\%$ and $58.68\%$, respectively. Compared with the baseline, the proposed GLGAN achieved great improvements in all performance metrics. Specifically, the OA, mF1 and mIoU attained by GLGAN (Swin-Base) were $81.97\%$, $78.01\%$ and $65.04\%$, respectively, which amounts to an improvement of $4.34\%$, $5.50\%$ and $6.36\%$ as compared to Advent. In addition, GLGAN (Swin-Large) attained higher scores in OA, mF1 and mIoU, by employing a more complex encoder. Among all the unsupervised methods examined, the proposed De-GLGAN achieved the highest scores on all overall matrices with the smallest performance gap as compared to the “Supervised” non-DA method. It is easy to observe in Table~\ref{tab:results1} that the proposed De-GLGAN attained the highest F1 score in four main categories, including $93.94\%$ for {\em Building}, $70.06\%$ for {\em Low Vegetation}, $70.40\%$ for {\em Cars} and $86.13\%$ for {\em Impervious Surface}, which amounts to improvement of $3.09\%$ for {\em Building}, $11.75\%$ for {\em Low Vegetation}, $15.33\%$ for {\em Cars} and $6.6\%$ for {\em Impervious Surface}. Upon comparing the results of GLGAN (Swin-Base) and De-GLGAN (Swin-Base), it becomes evident that improvements are observed across all categories, suggesting that the HLFD module facilitates the learning of transferable features.
Finally, Fig.~\ref{fig5} presents the semantic segmentation results achieved by part of the comparative methods. Visual inspection of Fig.~\ref{fig5} reveals that the proposed GLGAN and De-GLGAN are more accurate in differentiating visually similar categories such as {\em Trees} and {\em Low Vegetation}. Furthermore, the segmentation boundary of ground objects provided by our method is much smoother, with fewer random prediction points than other UDA methods.

\begin{table*}[t]\footnotesize
	\centering
	\caption{Task 2: Quantitative comparison of F1 (\%) on the adaptation of P-RGB to V-IRRG.}
	\setlength{\tabcolsep}{5mm}{
		\begin{tabular}{m{2.7cm}<{\centering}|m{1.4cm}<{\centering}|m{0.4cm}<{\centering}m{0.4cm}<{\centering}m{0.4cm}<{\centering}m{0.4cm}<{\centering}m{0.4cm}<{\centering}|m{0.4cm}<{\centering}m{0.4cm}<{\centering}m{0.4cm}<{\centering}}
			\hline
			\textbf{Method}   & \textbf{Backbone} & \textbf{Bui.} & \textbf{Tre.}  & \textbf{Low.}  & \textbf{Car}   & \textbf{Imp.}  & \textbf{OA}    & \textbf{mF1}    & \textbf{mIoU}   \\
			\hline
			Non-DA (Source-Only)   &   ResUNet++     & 64.48 &	0.04 & 5.26 & 1.57 & 58.21 & 35.71 & 25.91 & 18.43 \\
   		\hline
			DANN          &   -    & 73.20 &	62.46 &	41.50 &	54.94 &	64.39 &	61.85 &	59.30 &	42.94  \\
			AdasegNet     &   ResNet101    &	74.04 &	75.16 &	40.85 &	57.02 &	62.12 &	64.75 &	61.84 &	45.92 \\
			Advent        &   ResNet101    &	83.29 &	\textbf{77.85} &	42.37 &	49.92 &	65.73 &	67.26 &	63.83 &	48.84 \\
			GANAI         &   ResNet101    & 76.87 &	56.25 &	41.50 &	57.93 &	66.07 &	61.81 &	59.72 &	43.57 \\
			TriADA        &   ResNet101    & 77.86 &	73.85 &	36.57 &	58.70 &	66.48 &	66.09 &	62.69 &	47.20 \\
			CCAGAN        &   ResNet101    & 85.60 &	75.27 &	36.31 &	63.24 &	66.03 &	68.38 &	65.25 &	50.58 \\
      MBATA-GAN     &   ResNet101    & 87.67 &	75.30 &	39.06 &	64.95 &	67.33 &	69.29 & 66.86 & 52.31 \\
			MASNet        &   ResNet101    & 85.31 &	77.79 &	39.69 &	51.92 &	74.44 &	71.59 &	65.83 &	51.43 \\
			DSSFNet       &   ResNet101    & 87.15 &	77.66 &	48.77 &	56.02 &	73.46 &	73.14 &	68.61 &	53.98 \\
      \hline
			\multirow{2}{*}{GLGAN} &  Swin-Base &	87.71 &	68.37 &	58.59 & 66.68 & 73.73 &	73.21 & 71.02 & 55.98 \\
             &  Swin-Large &	89.70 &	77.53 &	58.93 & 63.21 & 78.48 &	77.23 & 73.57 & 59.44 \\
      De-GLGAN &  Swin-Base & \textbf{92.42} & 77.53 & \textbf{62.88} & \textbf{69.38} & \textbf{80.42} & \textbf{79.16} & \textbf{76.53} & \textbf{63.09} \\
			\hline
			Non-DA (Supervised)     &   ResUNet++    &	94.84 &	90.15 &	78.25 &	80.16 &	90.47 &	89.66 &	86.77 & 77.20 \\
			\hline
	\end{tabular}}\label{tab:results2}
\end{table*}

\begin{figure}[t]
	\centering
	{\includegraphics[width=0.95\linewidth]{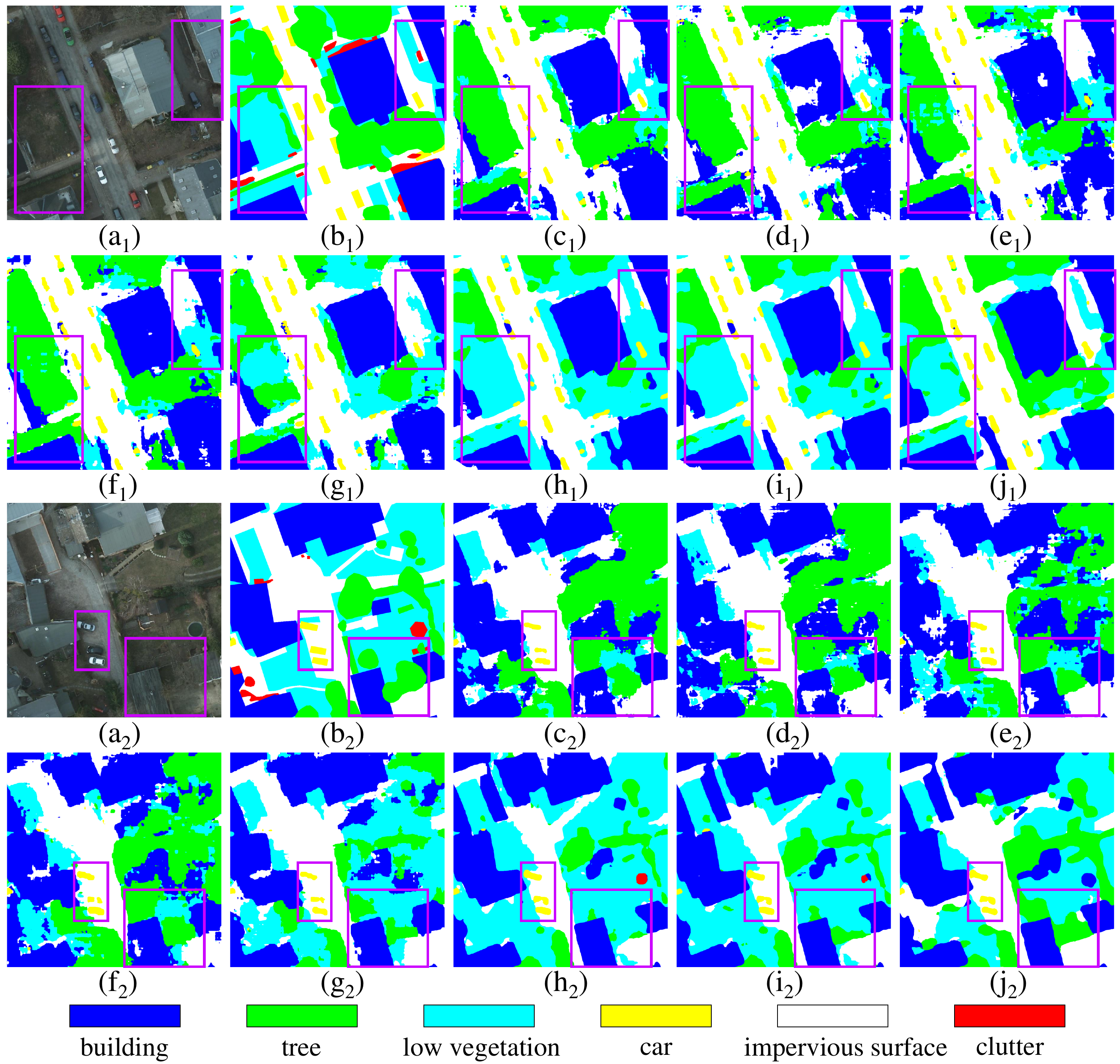}}
	\caption{Task 2: qualitative results of the adaptation P-RGB to V-IRRG with the size of $1024 \times 1024$, where the two samples are distinguished by subscripts. (a) IRRG images, (b) Ground Truth, (c) AdasegNet, (d) Advent, (e) TriADA, (f) CCAGAN, (g) MBATA-GAN (h) MASNet, (i) DSSFNet and (j) De-GLGAN. Purple boxes are marked to highlight the differences.}
	\label{fig6}
\end{figure}

\noindent{\emph{\textbf{Task 2} (P-RGB to V-IRRG):}}
Table~\ref{tab:results2} shows the results for Task $2$ that encounters a more severe domain shift problem than Task $1$. In addition to the discrepancies in the spatial resolution, the geographical location and other environmental factors, Task $2$ also has to learn image representations from the RGB bands before transferring the learned features to the IRRG bands. Notably, the ``Source-only" method showed the worst performance, which is evidenced by its poor OA, mF1, and mIoU of $35.71\%$, $25.91\%$ and $18.43\%$, respectively. In sharp contrast, it is observed from Table~\ref{tab:results2} that the GLGAN (Swin-Large) achieved OA of $77.23\%$, mF1 of $73.57\%$, and mIoU of $59.44\%$, which stands for an increase of $9.97\%$, $9.74\%$, and $10.60\%$ as compared to the corresponding performance of Advent, respectively. Further, the proposed De-GLGAN (Swin-Base) achieved a higher performance than the GLGAN (Swin-Large). It proved the broad applicability of the proposed HLFD, which can steadily improve the performance of the original UDA method in different scenarios. Moreover, these results demonstrate that the proposed GLGAN and De-GLGAN can effectively handle large domain shifts. Furthermore, it is interesting to observe that the De-GLGAN improved the accuracy for {\em Low Vegetation} by $20.51\%$, but suffered from performance degradation of $0.32\%$ for {\em Tree}. This observation could be explained by the fact that the trees in Potsdam are mostly sparse deciduous trees, while those in Vaihingen have denser foliage and canopy, as shown in Fig.~\ref{fig5_0}. As a result, {\em Tree} and {\em Low Vegetation} in Potsdam share similar spectral characteristics with {\em Low Vegetation} in Vaihingen. Finally, Fig.~\ref{fig6} illustrates some segmentation examples obtained in Task $2$ by various methods. It is clearly observed that the ``Source-only" method suffered from the worst performance. Furthermore, despite the fact that existing UDA methods showed improvements of different degrees, they incurred deficiencies in the aspects of providing clear boundaries and differentiating visually similar categories. By leveraging its powerful global-local feature extraction capability and frequency decomposition frequency technology, the proposed De-GLGAN achieved segmentation performance close to that of the supervised non-DA method.

\begin{table*}[t]\footnotesize
	\centering
	\caption{Task 3: Quantitative comparison of F1 (\%) on the adaptation of V-IRRG to P-RGB.}
	\setlength{\tabcolsep}{5mm}{
		\begin{tabular}{m{2.7cm}<{\centering}|m{1.4cm}<{\centering}|m{0.4cm}<{\centering}m{0.4cm}<{\centering}m{0.4cm}<{\centering}m{0.4cm}<{\centering}m{0.4cm}<{\centering}|m{0.4cm}<{\centering}m{0.4cm}<{\centering}m{0.4cm}<{\centering}}
			\hline
			\textbf{Method}     & \textbf{Backbone} & \textbf{Bui.} & \textbf{Tre.}  & \textbf{Low.}  & \textbf{Car}   & \textbf{Imp.}  & \textbf{OA}    & \textbf{mF1}    & \textbf{mIoU}   \\
			\hline
			Non-DA (Source-Only)    &   ResUNet++   & 42.87 &	0.10 & 7.36 &	3.55 & 43.27 & 34.14 & 19.43 & 12.11  \\
   		\hline
			DANN           &  -    & 63.70 &	42.50 &	40.26 &	58.21 &	62.10 &	54.53 &	53.35 &	37.00  \\
			AdasegNet      &  ResNet101    & 73.23 & 46.23 & 33.88 & 59.48 & 71.61 & 58.38 & 56.89 & 41.27 \\
			Advent         &  ResNet101    & 71.13 &	46.62 &	31.70 &	66.76 &	70.95 &	57.30 &	57.43 &	41.90 \\
			GANAI          &  ResNet101    &	64.97 &	38.79 &	45.29 &	67.45 &	67.17 &	57.02 &	56.73 &	40.58 \\
			TriADA         &  ResNet101    &	73.95 &	38.06 &	47.87 & 70.46 &	72.55 &	61.34 &	60.58 &	44.99 \\
			CCAGAN         &  ResNet101    &	76.45 &	44.86 &	47.28 &	72.82 &	71.26 &	61.68 &	62.53 &	46.87 \\
      MBATA-GAN      &  ResNet101    &  81.54 & 30.75 & 58.14 & 69.83 & 75.37 & 65.82 & 63.13 & 48.42 \\
			MASNet         &  ResNet101    & 77.89 & 44.87 & 61.89 & 76.04 & 73.97 & 66.42 & 66.93 & 51.51 \\
			DSSFNet        &  ResNet101    & 79.82 & 50.62 & 64.68 & 76.86 & 75.47 & 68.87 & 69.49 & 54.22 \\
      \hline
			\multirow{2}{*}{GLGAN} & Swin-Base	& 82.77 & 60.46 & 65.19 & 77.27 & 76.14 & 71.15 & 72.37 & 57.34 \\
             & Swin-Large &	\textbf{85.65} &	62.08 &	69.26 &  \textbf{81.37} & 76.09 & 73.63 & 74.89 & 60.58 \\
      De-GLGAN      & Swin-Base	& \textbf{85.65} & \textbf{65.61} & \textbf{71.02} & 79.43 & \textbf{79.66} & \textbf{75.54} & \textbf{76.27} & \textbf{62.17} \\
			\hline
			Non-DA (Supervised)    &   ResUNet++    & 92.08 & 74.91 & 80.13 & 90.79 & 89.35 & 84.69 & 85.46 & 75.19 \\
			\hline
	\end{tabular}}\label{tab:results3}
\end{table*}

\begin{figure}[h]
	\centering
	{\includegraphics[width=0.95\linewidth]{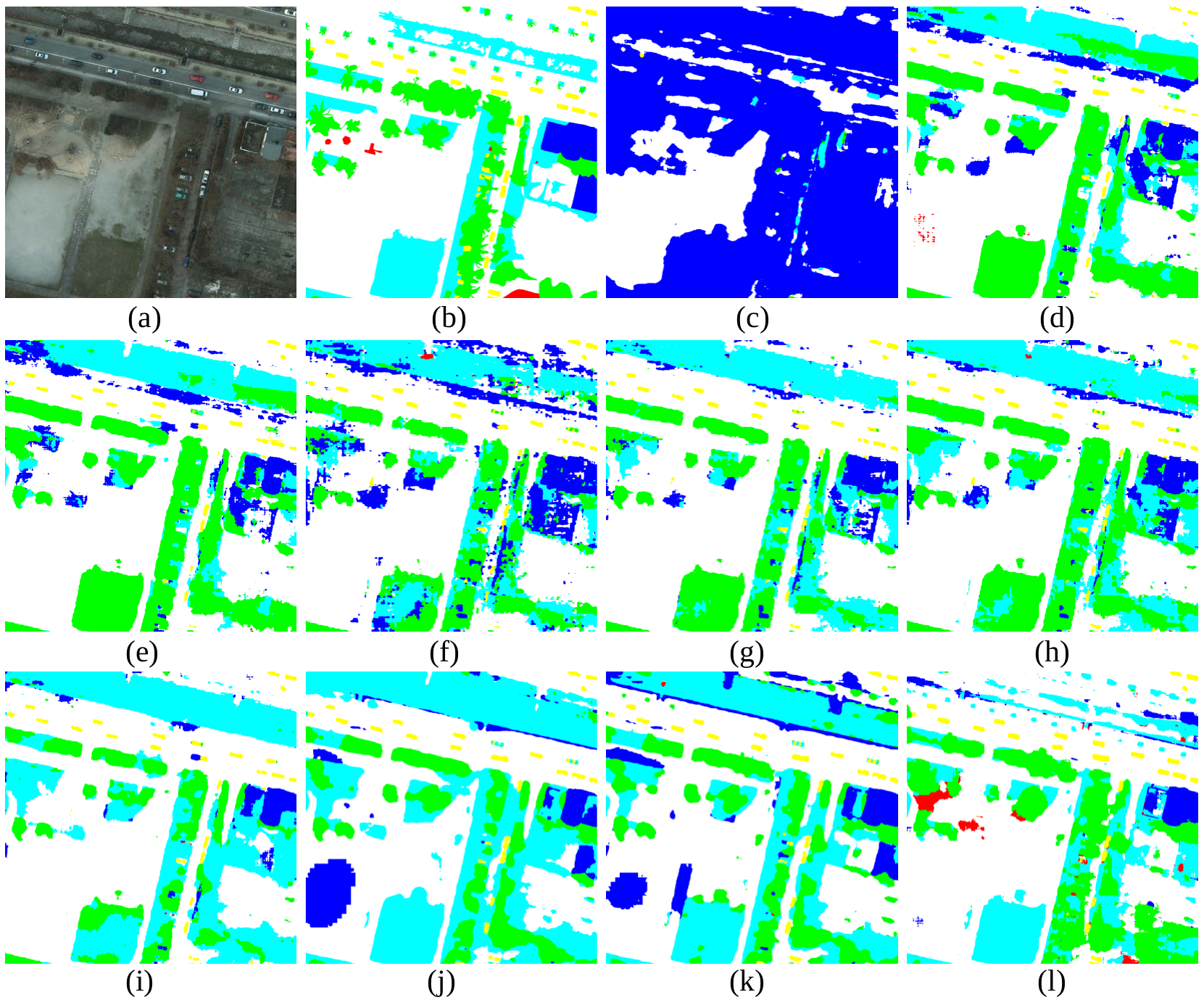}}
	\caption{Task 3: qualitative results of the adaptation V-IRRG to P-RGB with the size of $1024 \times 1024$, where the two samples are distinguished by subscripts. (a) RGB images, (b) Ground Truth, (c) AdasegNet, (d) Advent, (e) TriADA, (f) CCAGAN, (g) MBATA-GAN (h) MASNet, (i) DSSFNet and (j) De-GLGAN. Purple boxes are marked to highlight the differences.}
	\label{fig7}
\end{figure}

\noindent{\emph{\textbf{Task 3} (V-IRRG to P-RGB):}}
Task $3$ suffers from a domain shift problem similar to Task $2$, except that its source domain has a smaller data size than the target domain. From Table~\ref{tab:results3}, it is observed that De-GLGAN achieved the highest OA of $75.54\%$, mF1 of $76.27\%$, and mIoU of $62.17\%$, which stands for an improvement of $18.24\%$, $18.84\%$, and $20.27\%$ as compared to Advent, respectively. Furthermore, the segmentation accuracy on {\em Tree} and {\em Low vegetation} in Task $3$ has been substantially improved as compared to Task $2$. This is because these two categories in Vaihingen show much different spectral characteristics, providing more transferable features. Some segmentation samples are presented in Fig.~\ref{fig7} for visual inspection. In sharp contrast to other UDA methods, the proposed methods provided the most accurate semantic information with the smoothest boundaries.

\begin{figure}[t]
	\centering
{\includegraphics[width=0.9\linewidth]{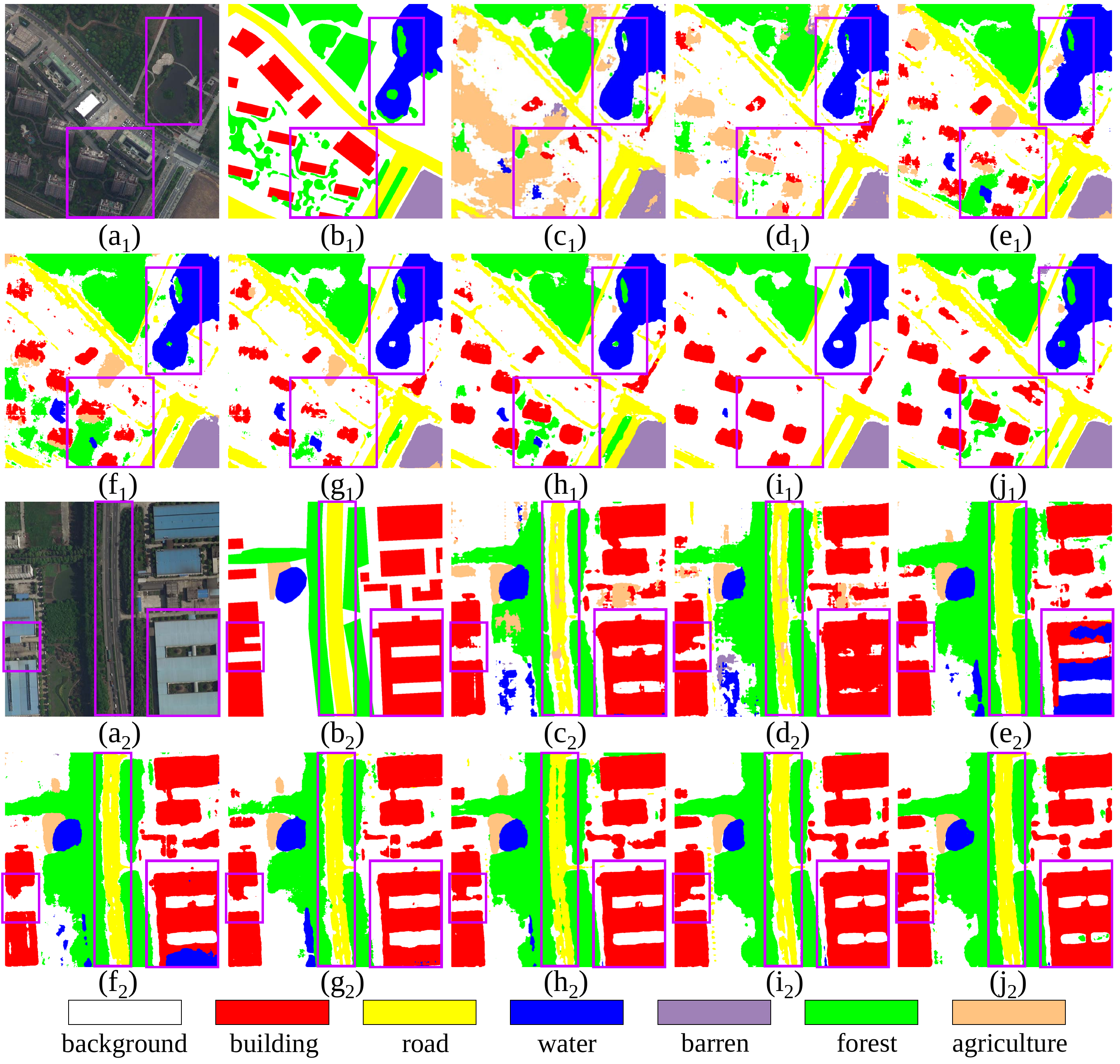}}
	\caption{Task 4: qualitative results of the adaptation Rural to Urban with the size of $1024 \times 1024$, where the two samples are distinguished by subscripts. (a) RGB images, (b) Ground Truth, (c) AdasegNet, (d) Advent, (e) TriADA, (f) CCAGAN, (g) MBATA-GAN (h) MASNet, (i) DSSFNet and (j) De-GLGAN. Purple boxes are marked to highlight the differences.}
	\label{fig7_r2u}
\end{figure}

\noindent{\emph{\textbf{Task 4} (Rural to Urban):}}
Inspection of Table~\ref{tab:results4} suggests that the proposed De-GLGAN achieved the highest OA, mF1 and mIoU among all the unsupervised methods. Compared with the baseline, the proposed De-GLGAN achieved great improvements in all performance metrics. Specifically, the OA, mF1 and mIoU attained were $70.01\%$, $70.93\%$ and $55.50\%$, respectively, which amounts to an improvement of $6.33\%$, $6.32\%$ and $7.4\%$ as compared to Advent. Notably, the proposed methods attained great performance in two categories, including $59.53\%$ for {\em Background} and $74.49\%$ for {\em Building}. These two categories account for more than half of the distribution of Urban scene as presented in Fig.~\ref{fig5_1}(c). It revealed that our approaches effectively learn the transferable features of the dominant object categories in the rural scene, which can aid in identifying these categories in the urban scene. Fig.~\ref{fig7_r2u} shows the semantic segmentation results for this task. Visual inspection of Fig.~\ref{fig7_r2u} reveals that our method still has fewer random prediction points, maintaining the integrity of the ground objects.

\begin{table*}[t]\scriptsize
	\centering
	\caption{Task 4: Quantitative comparison of F1 (\%) on the adaptation of Rural to Urban.}
	\setlength{\tabcolsep}{5mm}{
		\begin{tabular}{m{2.0cm}<{\centering}|m{1.2cm}<{\centering}|m{0.24cm}<{\centering}m{0.24cm}<{\centering}m{0.24cm}<{\centering}m{0.24cm}<{\centering}m{0.24cm}<{\centering}m{0.24cm}<{\centering}m{0.24cm}<{\centering}|m{0.3cm}<{\centering}m{0.3cm}<{\centering}m{0.3cm}<{\centering}}
			\hline
			\textbf{Method}     & \textbf{Backbone} & \textbf{Bac.} & \textbf{Bui.}  & \textbf{Road}  & \textbf{Wat.}   & \textbf{Bar.} & \textbf{For.} & \textbf{Agr.}  & \textbf{OA}    & \textbf{mF1}    & \textbf{mIoU}   \\
			\hline
			DANN    &  -   & 53.38 & 60.11 & 56.59 & 74.63 & 58.51 & 64.80 & 65.19 & 61.71 & 61.89 & 45.14 \\
			AdasegNet    &  ResNet101   & 54.76 & 62.77 & 60.84 & 80.75 & 62.28 & 63.23 & 65.57 & 63.02 & 64.31 & 47.87 \\
			Advent       &  ResNet101   & 56.40 & 65.53 & 62.52 & 77.98 & 57.72 & 65.53 & 66.60 & 63.68 & 64.61 & 48.09 \\
			GANAI        &  ResNet101   & 55.54 & 61.04 & 54.76 & 75.42 & 65.31 & 65.01 & 67.31 & 62.98 & 63.48 & 46.86 \\
			TriADA       &  ResNet101   & 57.05 & 63.35 & 65.17 & 80.56 & \textbf{65.36} & 67.16 & 71.53 & 66.24 & 67.17 & 50.97 \\
			CCAGAN       &  ResNet101   & 56.91 & 67.45 & 65.68 & 83.00 & 65.10 & 65.93 & 74.93 & 67.78 & 68.43 & 52.55 \\
			MBATA-GAN    &  ResNet101   & 58.66 & 64.95 & 65.90 & 82.54 & 64.35 & 67.72 & 74.18 & 67.93 & 68.47 & 52.52 \\
			MASNet       &  ResNet101   & 59.49 & 71.28 & 64.15 & 83.36 & 63.34 & 69.97 & \textbf{75.99} & 69.52 & 69.65 & 53.98 \\
			DSSFNet      &  ResNet101   & 59.25 & 73.95 & 65.84 & 84.44 & 64.12 & \textbf{70.74} & 72.89 & 68.90 & 70.18 & 54.60 \\
			\hline
			\multirow{2}{*}{GLGAN} & Swin-Base  & 58.06 & 73.11 & \textbf{69.99} & 84.09 & 59.70 & 70.66 & 73.64 & 69.13 & 69.89 & 54.32 \\
             & Swin-Large &	57.83 & 74.45 & 67.90 & \textbf{84.57} & 63.08 & 69.09 & 75.68 & 69.50 & 70.37 & 54.80 \\
			De-GLGAN  & Swin-Base  & \textbf{59.53} & \textbf{74.49} & 68.80 & 84.66 & 63.71 & 70.50 & 74.83 & \textbf{70.01} & \textbf{70.93} & \textbf{55.50} \\
			\hline
	\end{tabular}}\label{tab:results4}
\end{table*}

\begin{table*}[t]\scriptsize
	\centering
	\caption{Task 5: Quantitative comparison of F1 (\%) on the adaptation of Urban to Rural.}
	\setlength{\tabcolsep}{5mm}{
		\begin{tabular}{m{2.0cm}<{\centering}|m{1.2cm}<{\centering}|m{0.24cm}<{\centering}m{0.24cm}<{\centering}m{0.24cm}<{\centering}m{0.24cm}<{\centering}m{0.24cm}<{\centering}m{0.24cm}<{\centering}m{0.24cm}<{\centering}|m{0.3cm}<{\centering}m{0.3cm}<{\centering}m{0.3cm}<{\centering}}
			\hline
			\textbf{Method}     & \textbf{Backbone} & \textbf{Bac.} & \textbf{Bui.}  & \textbf{Road}  & \textbf{Wat.}   & \textbf{Bar.} & \textbf{For.} & \textbf{Agr.}  & \textbf{OA}    & \textbf{mF1}    & \textbf{mIoU}   \\
			\hline
			DANN    &  -   & 67.80 & 49.19 & 24.80 & 57.99 & 26.04 & 7.30 & 28.43 & 55.04 & 37.36 & 24.89 \\
			AdasegNet    &  ResNet101   & 66.47 & 62.36 & 34.59 & 68.49 & 27.85 & 15.05 & 17.04 & 54.85 & 41.69 & 28.82 \\
			Advent       &  ResNet101   & 68.49 & 62.01 & 32.89 & 54.88 & \textbf{33.65} & 5.44 & 39.92 & 57.82 & 42.47 & 28.93 \\
			GANAI        &  ResNet101   & 67.75 & 63.08 & 30.20 & \textbf{72.08} & 20.00 & 7.62 & 26.21 & 56.64 & 40.99 & 28.80 \\
			TriADA       &  ResNet101   & 69.91 & 66.06 & \textbf{53.17} & 57.67 & 18.76 & 22.61 & 54.99 & 62.58 & 49.02 & 34.40 \\
			CCAGAN       &  ResNet101   & 70.34 & 65.28 & 45.76 & 68.60 & 21.97 & 11.72 & 55.33 & 63.45 & 48.43 & 34.48 \\
			MBATA-GAN    &  ResNet101   & 70.00 & 64.77 & 49.93 & 67.46 & 16.03 & 7.19 & 51.53 & 62.12 & 46.70 & 33.27 \\
			MASNet       &  ResNet101   & 68.71 & 67.13 & 41.53 & 55.06 & 19.11 & 12.10 & 49.46 & 60.18 & 44.73 & 30.99 \\
			DSSFNet      &  ResNet101   & 70.88 & 61.88 & 45.79 & 65.88 & 23.94 & 29.88 & 52.54 & 63.01 & 50.11 & 35.01 \\
			\hline  
			\multirow{2}{*}{GLGAN} & Swin-Base  &	69.68 & 63.65 & 40.80 & 69.87 & 25.72 & 6.65 & 44.15 & 60.90 & 45.79 & 32.27 \\
             & Swin-Large &	69.74 & 67.80 & 51.73 & 71.91 & 15.10 & 8.20 & 48.63 & 62.20 & 47.59 & 34.34 \\
			De-GLGAN  & Swin-Base  & \textbf{71.48} & \textbf{70.96} & 49.92 & 69.12 & 23.79 & \textbf{33.60} & \textbf{56.83} & \textbf{65.22} & \textbf{53.67} & \textbf{38.58} \\
			\hline
	\end{tabular}}\label{tab:results5}
\end{table*}

\begin{figure}[t]
	\centering
{\includegraphics[width=0.9\linewidth]{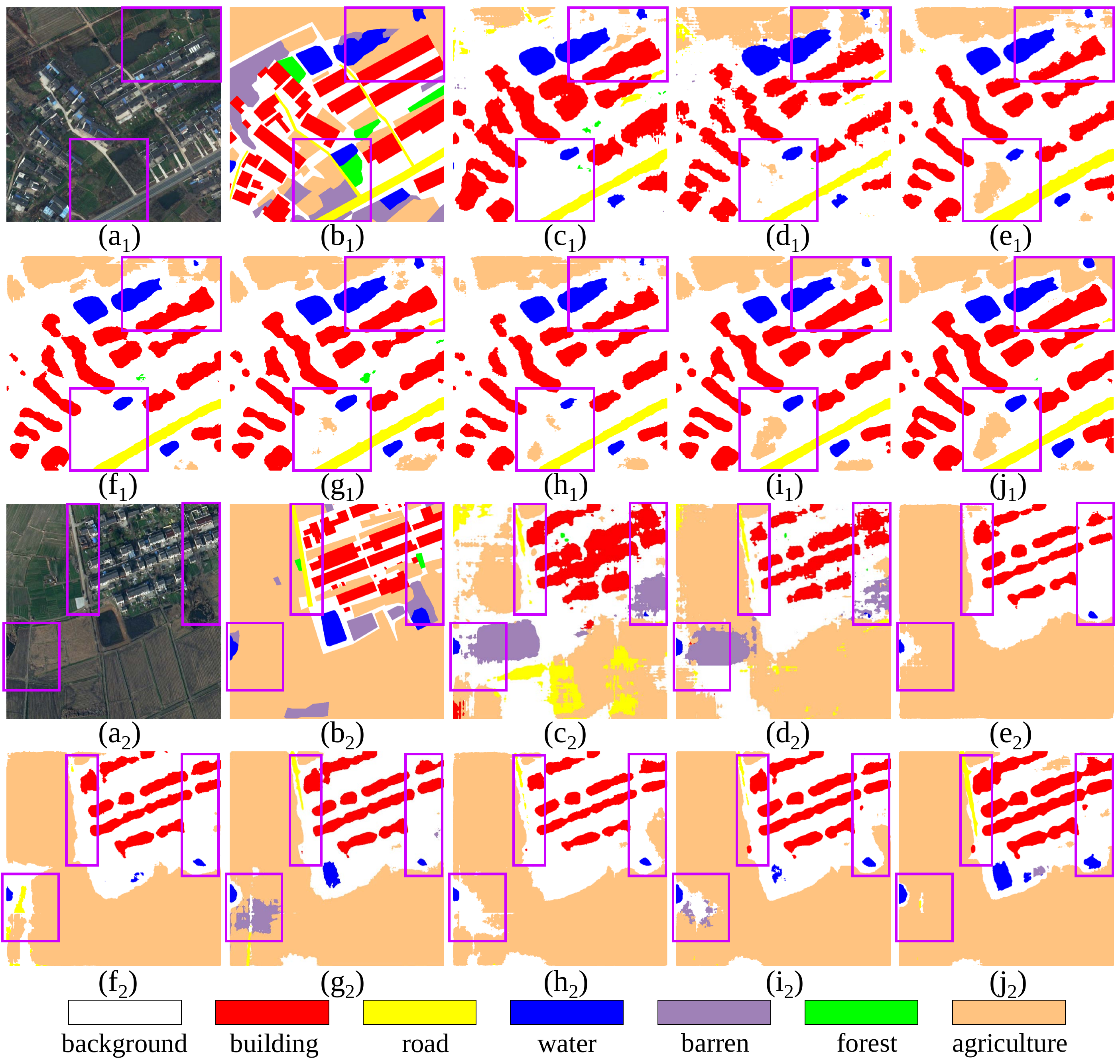}}
	\caption{Task 5: qualitative results of the adaptation Urban to Rural with the size of $1024 \times 1024$, where the two samples are distinguished by subscripts. (a) RGB images, (b) Ground Truth, (c) AdasegNet, (d) Advent, (e) TriADA, (f) CCAGAN, (g) MBATA-GAN (h) MASNet, (i) DSSFNet and (j) De-GLGAN. Purple boxes are marked to highlight the differences.}
	\label{fig7_u2r}
\end{figure}

\noindent{\emph{\textbf{Task 5} (Urban to Rural):}}
Compared with Task $4$, Task $5$ is more challenging since the minority categories in the Urban scene, including {\em Water}, {\em Forest} and {\em Agriculture}, are precisely the categories of objects with complex characteristics. Specifically, compared with objects such as {\em Building} and {\em Road}, they have more complicated and variable boundaries and spatial relationships. From Table~\ref{tab:results5}, it is observed that the proposed De-GLGAN achieved the highest OA of $65.22\%$, mF1 of $53.67\%$, and mIoU of $38.58\%$, which stands for an improvement of $7.4\%$, $11.2\%$, and $9.65\%$ as compared to Advent, respectively. Furthermore, the segmentation accuracy on {\em Building}, {\em Forest} and {\em Agriculture} in Task $5$ has been greatly improved. Observing the results of De-GLGAN and GLGAN, we can see that the frequency decomposition technique has a powerful feature extraction capability for this difficult category. By comparing the results of GLGAN and De-GLGAN, it is evident that the frequency decomposition technique excels in feature extraction for the challenging categories. The HLFD module can focus on both local details and global semantics, enabling the extraction of transferable features from a minimal number of category samples.
The visualized segmentation samples shown in Fig.~\ref{fig7_u2r} indicated that De-GLGAN was more accurate in predicting {\em Agriculture} and {\em Forest}. It further demonstrates that our method can effectively learn transferable features from a limited number of samples.

The comparative results reveal a performance gap between our method and other methods. This is because the proposed framework improves both the generator and discriminator of the GAN framework. In summary, the extensive experiments demonstrated that the proposed GLGAN and De-GLGAN have the capability of overcoming domain shifts, leading to significantly improved semantic segmentation performance on different remote sensing scenarios.

\subsection{Ablation Studies}\label{sec:abla}
Given that the proposed De-GLGAN incorporates both GLTB and HLFD, we conducted three sets of ablation studies to demonstrate the importance of each component.

\noindent{\emph{1) Global-Local information:}}
To verify the effectiveness of the proposed backbone GLGAN, eight ablation experiments were performed as shown in Table~\ref{tab:abla1}. The baseline shown in the first row was obtained with a purely convolutional model composed of an encoder (ResNet101), a decoder (ASPP) and a CNN-based discriminator, whereas the second row shows the case for the baseline with ResNet101 being replaced by the Swin Transformer for feature extraction. Furthermore, the third row shows the case in which the ASPP in the baseline was replaced by the GLTB-based decoder to recover the semantic features and predict the segmentation maps, while the CNN-based discriminator in the baseline was substituted with GLDis in the fourth row. The next three rows, i.e., the fifth row to the seventh row, are pairwise combinations of the Swin Transformer-based encoder, the GLTB-based decoder and GLDis.
Inspection of the first four rows in Table~\ref{tab:abla1} suggests that each proposed component is of great necessity in the proposed GLGAN. Notably, the result in the second row shows that Swin Transformer only contributes a small part of the superior performance of GLGAN. Furthermore, the ablation experiments shown in the second and fifth rows confirmed the importance of the decoder in learning a transferable output space collocated with the Swin Transformer. Comparison of the results reported in the second and sixth rows, the third and seventh rows, and the fifth and last rows suggest that GLDis was able to improve overall segmentation performance, which provides strong evidence to support our motivation to strengthen the discriminator design under the GAN framework. Therefore, the outstanding performance of GLGAN benefits from the combined effect of all the aforementioned modules. Clearly, GLDis is a better choice than the conventional CNN-based discriminator. Table~\ref{tab:abla1} confirmed that the proposed GLGAN is more efficient in image feature extraction and semantic information recovery by effectively exploiting intra- and inter-domain features.

\begin{table}[h]\scriptsize
\begin{threeparttable}
	\centering
	\caption{Ablation studies on the structure of GLGAN (\%). Bold values are the best.}
	\setlength{\tabcolsep}{5mm}{
		\begin{tabular}{m{0.5cm}<{\centering}m{0.5cm}<{\centering}m{0.5cm}<{\centering}|m{0.4cm}<{\centering}m{0.4cm}<{\centering}m{0.4cm}<{\centering}}
			\hline
                \multicolumn{2}{c}{\textbf{Generator}} & \multirow{2}{*}{\textbf{GLDis}} & \multirow{2}{*}{\textbf{OA}}   & \multirow{2}{*}{\textbf{mF1}}  & \multirow{2}{*}{\textbf{mIoU}}\\ 
                \textbf{encoder} & \textbf{decoder}  &     &     &    &  \\
			\hline
                &    &            & 77.63 &	72.51 &	58.68 \\
                \hline
                \checkmark &  &    & 80.42 & 73.83 & 60.32 \\
                &  \checkmark  &    & 77.61 & 74.15 & 60.23 \\
                &    & \checkmark  & 77.46 & 73.71 & 59.71 \\
                \checkmark &  \checkmark &  & 80.51 & 77.27 & 64.03 \\
                \checkmark &  & \checkmark  & 81.63 & 74.40 & 61.28 \\
                & \checkmark   & \checkmark  & 78.23 & 75.17 & 61.26 \\
                \hline
                \checkmark & \checkmark  &  \checkmark   & \textbf{81.97} & \textbf{78.01} & \textbf{65.04}  \\
			\hline
	\end{tabular}}\label{tab:abla1}
\end{threeparttable}
\end{table}

\begin{table}[h]\centering
	\begin{threeparttable}
		\centering
		\caption{Ablation studies for high/low-frequency of De-GLGAN (\%). Bold values are the best.}
		\setlength{\tabcolsep}{5mm}{
			\begin{tabular}{cc|ccc}
				\hline
				\textbf{HF} &\textbf{LF}  &   \textbf{OA}  &  \textbf{mF1}  & \textbf{mIoU} \\
				\hline
				&            & 81.97 & 78.01 & 65.04 \\
				\hline
				\checkmark& &  83.40 & 79.63 & 67.21 \\
				& \checkmark&  83.38 & 79.79 & 67.45 \\
				\hline
				\checkmark& \checkmark & \textbf{83.66} & \textbf{80.30} & \textbf{68.09}  \\
				\hline
		\end{tabular}}\label{tab:abla2}
	\end{threeparttable}
\end{table}
\noindent{\emph{2) High/Low-frequency information:}}
To verify the effectiveness of the two branches in HLFD, four ablation experiments were performed, as shown in Table~\ref{tab:abla2}. The baseline shown in the first row was the results provided by GLGAN. The second row shows the case in which only the high-frequency extractor is employed to align the high-frequency information, whereas the third row shows the case in which only the low-frequency extractor is employed to align the low-frequency information. The last row presents the results achieved by utilizing the complete HLFD. Inspection of the results suggests that in UDA tasks, the separate alignment of high- and low-frequency information is very important when learning cross-domain transferable features. We can also draw an inference that besides sharing similarities in global features, the source and target domains also exhibit resemblances in local features, such as the shape of tree crowns and vehicle outlines. These local characteristics of ground objects can be analyzed from a frequency perspective, thereby offering valuable insights for unlabeled target domains.

\noindent{\emph{3) Multiscale features:}}
Eight ablation experiments were performed, as shown in Table~\ref{tab:abla3} to verify the necessity of aligning multiscale features. The baseline shown in the first row still included the results provided by GLGAN. According to the experimental results we obtained, we present the ablation experiments in the order shown in Table~\ref{tab:abla3}. Specifically, we investigated the impact of a single scale, ranging from shallow features (scale 1) to deep features (scale 4), before progressively incorporating additional scales. 
There are two interesting conclusions confirmed by Table~\ref{tab:abla3}. Firstly, the alignment of multiscale features is necessary. Secondly, deep features hold greater significance compared to shallow features, which has been also reported in the literature \citep{chen2022unsupervised, mbatagan, liu2022unsupervised}.
In summary, the proposed GLGAN and De-GLGAN achieved great segmentation performance as compared to other state-of-the-art UDA methods. Both quantitative results and visual inspection confirmed that GLTB-based backbone GLGAN and the frequency decomposition technology can largely bridge the segmentation performance gap between UDA methods and supervised methods.

\begin{table}[h]\scriptsize
\begin{threeparttable}
	\centering
	\caption{Ablation studies on the multiscale features of De-GLGAN (\%). Bold values are the best.}
	\setlength{\tabcolsep}{3mm}{
		\begin{tabular}{m{0.8cm}<{\centering}m{0.8cm}<{\centering}m{0.8cm}<{\centering}m{0.8cm}<{\centering}|m{0.4cm}<{\centering}m{0.4cm}<{\centering}m{0.4cm}<{\centering}}
			\hline
      \textbf{Scale 1} &\textbf{Scale 2} & \textbf{Scale 3}  &  \textbf{Scale 4}   &   \textbf{OA}  &  \textbf{mF1}  & \textbf{mIoU} \\
			\hline
			&&    &            & 81.97 & 78.01 & 65.04 \\
			\hline
			 \checkmark & & &   & 82.89 & 78.82 & 66.13 \\
			& \checkmark & &   & 83.06 & 78.83 & 66.26 \\
			& & \checkmark &   & 82.44 & 79.29 & 66.68 \\
		  & &  & \checkmark  & 83.34 & 79.45 & 66.96 \\
			& & \checkmark &  \checkmark  & 83.22 & 79.83 & 67.35 \\
			& \checkmark&  \checkmark & \checkmark & 83.39 & 80.17 & 67.92  \\
			\hline
			\checkmark& \checkmark & \checkmark &  \checkmark   & \textbf{83.66} & \textbf{80.30} & \textbf{68.09}  \\
			\hline
	\end{tabular}}\label{tab:abla3}
\end{threeparttable}
\end{table}

\subsection{Parameter Analysis}\label{sec:para}
\begin{figure}[h]
	\centering
	{\includegraphics[width=0.95\linewidth]{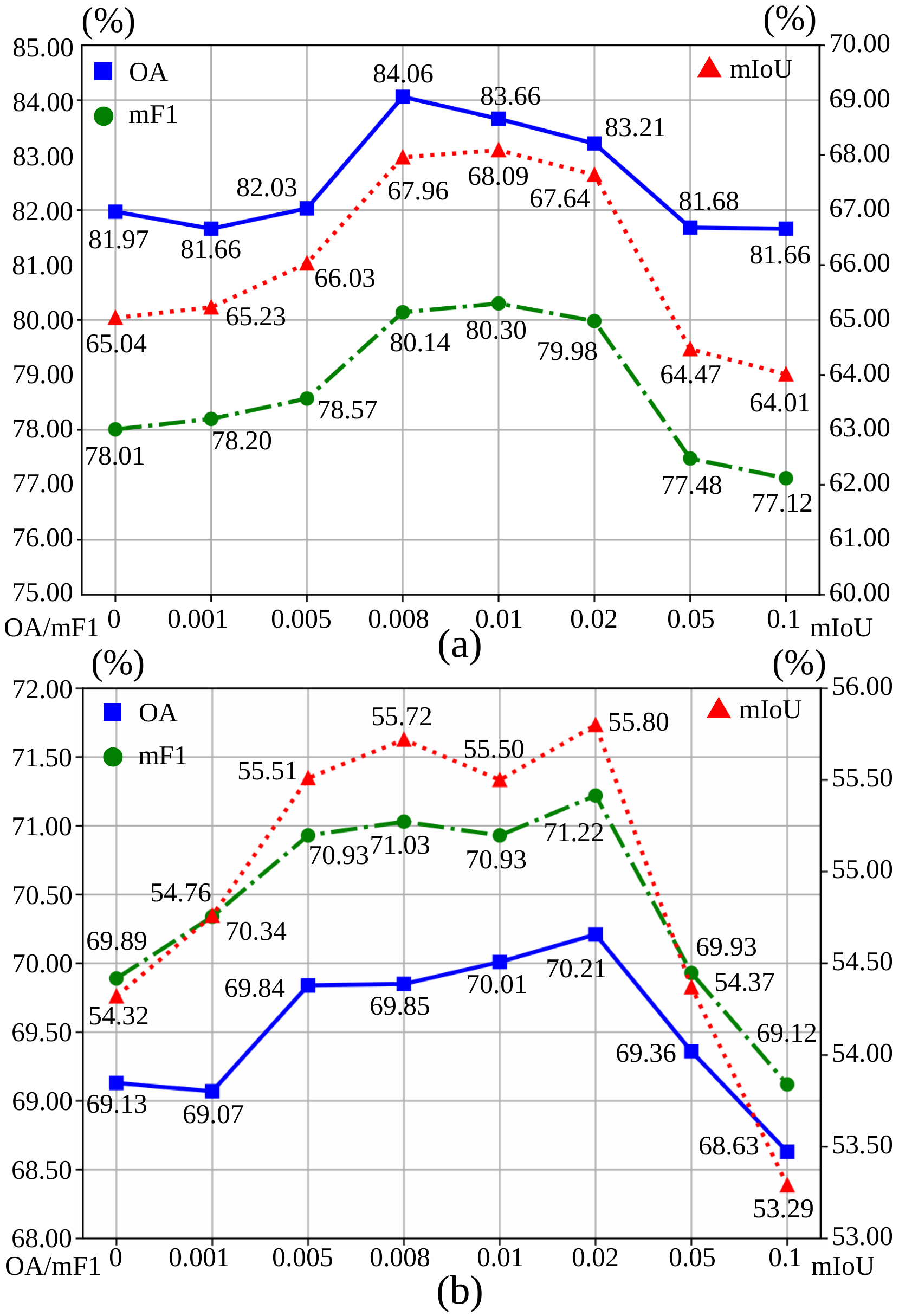}}
	\caption{Parameter analysis for high/low-frequency of De-GLGAN on (a) P-IRRG to V-IRRG and (b) Rural to Urban. The result indicates that as $\lambda^{f}$ increases, the impact of feature decomposition gradually improves until excessively large values lead to overemphasis.}
	\label{fig9}
\end{figure}

In this section, the sensitivity of hyper-parameter $\lambda^{f}$ is analyzed. This parameter is designed to regulate the contribution derived from frequency decomposition. Inspection of the results presented in Fig.~\ref{fig9} indicates that a smaller value of $\lambda^{f}$ diminishes the importance of aligning high/low-frequency information, whereas a larger value may exaggerate the significance of the frequency alignment. Consequently, setting $\lambda^{f}> 0.05$ results in notable performance degradation. Conversely, both performance demonstrates less sensitivity to $\lambda^{f}$ within the range $\lambda^{f}\in \left[0.008,0.02\right]$. Therefore, we set $\lambda^{f}=0.01$ in our experiments, which performs effectively across the five UDA tasks, suggesting its strong applicability.

\subsection{Visualization of Feature Decomposition}
\begin{figure}[h]
\centering
{\includegraphics[width=0.95\linewidth]{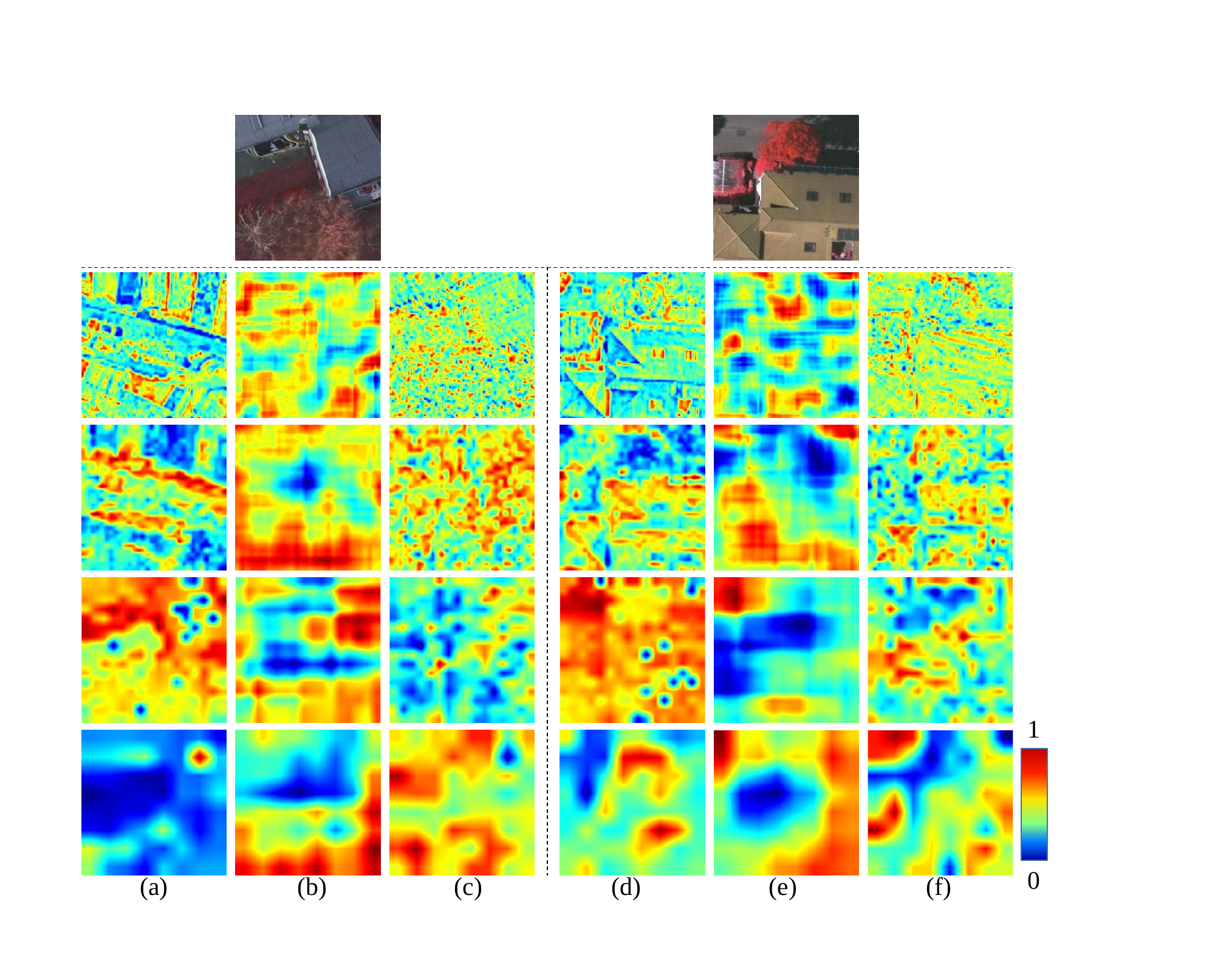}}
\caption{Visualization of feature maps before and after decomposition. The first row displays the input images from the source and the target domains. The second to the fifth row corresponds to the four scales. Source domain feature maps: (a) before decomposition, (b) low-frequency and (c) high-frequency. Target domain feature maps: (d) before decomposition, (e) low-frequency and (f) high-frequency. }
\label{fig8}
\end{figure}

In this section, we will investigate the characteristics of the proposed multiscale HLFD modules by visually inspecting the heatmaps before and after the frequency decomposition. Specifically, multiscale feature maps are first resized to a uniform image size, followed by averaging across all channels. Subsequently, pseudo-color processing is applied. The results are depicted in Fig.~\ref{fig8}. Upon inspecting the feature maps before and after the decomposition in both the source and target domains, distinctions become apparent between low- and high-frequency information: low-frequency features possess global characteristics, whereas high-frequency features appear more discrete and localized. In particular, even the fourth scale (high-level) feature maps show significant differences. It fully demonstrates that our HLFD modules successfully decompose high- and low-frequency information.

Comparison on the feature maps between the source and target domains reveals that cross-domain high-frequency features share great similarities as well as the cross-domain low-frequency features. It suggests that the model's capacity to decompose features acquired from the labeled source domain can be successfully applied to the target domain. In other words, the model has learned domain-invariant features from both high- and low-frequency, which correspond to the similarity of spatial details across domains that we emphasized in Sec.~\ref{sec:int} and the global context similarity that the existing works explored. Our work has proved that both of these components are necessary for cross-domain alignment in the UDA method.

\subsection{Model Complexity Analysis}\label{sec:complexity}
Table~\ref{tab:scale} shows the computational complexity of all UDA methods discussed. Compared with the baseline Advent, the proposed general backbone GLGAN (Swin-Base) achieved greatly improved segmentation performance with significantly reduced computational complexity as its hierarchical window-based attention has linear computation complexity to input image size \citep{liu2021swin}. More specifically, the computational complexity of GLGAN (Swin-Base) is only half that of Advent. Furthermore, GLGAN (Swin-Large) has computational complexity comparable to Advent. However, GLGAN (Swin-Large) requires much more memory to store its parameters. Overall, the proposed GLGAN harvests better performance at the cost of moderately increased computational complexity. The model complexity of De-GLGAN is between GLGAN (Swin-Base) and GLGAN (Swin-Large), yet it exhibits superior performance. It proved that the HLFD module is of great significance in practical application.

\begin{table}[htp]
	\centering
	\caption{Comparison on computational complexity measured by a $256 \times 256$ input on a single NVIDIA GeForce RTX 3090 GPU. Bold values are the best}
	\setlength{\tabcolsep}{0.1mm}{
		\begin{tabular}{m{1.8cm}<{\centering}|m{1.5cm}<{\centering}|m{1.5cm}<{\centering}m{1.5cm}<{\centering}m{1.2cm}<{\centering}m{1.2cm}<{\centering}}
			\hline
			\textbf{Model}   & \textbf{Backbone} & \textbf{Complexity (GFLOPs)}   &  \textbf{Parameters (M)} & \textbf{Memory (MB)}  &  \textbf{Speed (FPS)}  \\
			\hline
			ResUNet++      &  ResUNet++ &  241.82 	& 	\textbf{35.46} &     4365 	& 	25.10 	\\
			AdasegNet      &  ResNet101 &  47.59 	& 	42.83 	& 	2491 	& 	\textbf{43.31} 	\\
			Advent         &  ResNet101 &  47.95  & 	43.16 	& 	2491 	& 	30.65   \\
			GANAI          &  ResNet101 &  \textbf{10.27} & 		44.55 & 	\textbf{1075} 	& 	6.30 	  \\
			TriADA         &  ResNet101 &  22.15  & 	59.23 	& 	\textbf{1075} 	& 	4.54    \\
			CCAGAN         &  ResNet101 &  145.14 & 	104.02  & 	4503 	& 	13.04   \\
      MBATA-GAN      &  ResNet101 &  174.20 & 	143.90 	& 	5295 	& 	8.11 	  \\
			MASNet         &  ResNet101 &  82.34  &   64.84   &   3648  &   18.22   \\
			DSSFNet        &  ResNet101 &  128.52 &   112.74  &   4836  &   11.13   \\
		  \hline
			\multirow{2}{*}{GLGAN}    &  Swin-Base & 26.13  & 96.72 & 3639	& 12.50  \\
          &  Swin-Large & 45.97  & 159.03 & 4840	& 9.95  \\
			De-GLGAN  & Swin-Base & 34.26 & 138.68 & 3841 & 6.22 \\
			\hline
	\end{tabular}}\label{tab:scale}
\end{table}

\section{Conclusion}\label{sec:con}
In this work, a novel decomposition-based UDA approach for remote sensing image semantic segmentation has been proposed. Based on the observation that local detailed features and global contexts are both crucial for reducing domain discrepancy in semantic segmentation, a novel HLFD scheme has been proposed to conduct domain alignment in different subspaces through the decomposition of multiscale features from the frequency perspective. Furthermore, to exploit the global-local cross-domain dependencies, a fully GLGAN has been developed to learn robust domain-invariant representations, leveraging GLTBs. In particular, a novel discriminator, GLDis, has been established to enhance the generator's representation capability. Finally, a novel UDA framework called De-GLGAN has been established by integrating HLFD and GLGAN for cross-domain semantic segmentation. Extensive experiments on two UDA benchmarks, namely ISPRS Potsdam and Vaihingen, and LoveDA Rural and Urban, have confirmed that the proposed De-GLGAN can substantially outperform existing UDA methods while greatly bridging the performance gap between the existing adversarial training-based UDA approach and the fully-supervised training. Specifically, it can significantly improve the cross-domain transferability and generalization capability of the semantic segmentation model.

In future work, we will investigate the possibility of incorporating the decomposition-based feature learning strategy into fine-tuning visual foundation models on new semantic segmentation datasets. Meanwhile, it is interesting to investigate the integration of the proposed approach with reliable self-training to incorporate category-wise supervision more effectively. Furthermore, it is worth exploring the potential cross-domain applications of GLGAN and the decomposition-based alignment strategy to other tasks, such as image understanding \cite{rahnemoonfar2021floodnet, wang2024earthvqanet} and object detection \cite{cheng2023towards}.

\small
\bibliographystyle{IEEEtranN}
\bibliography{references}
\end{document}